\documentclass{article}

\usepackage{microtype}
\usepackage{graphicx}
\usepackage{wrapfig}
\usepackage{booktabs} 

\usepackage{hyperref}

\usepackage{soul}
\usepackage[font=footnotesize]{subfig}
\usepackage{tikz}
\usepackage{anyfontsize}
\usepackage{pifont}
\usepackage{diagbox}
\usepackage{nicematrix}
\usepackage[per-mode=symbol]{siunitx}
\usepackage{svg}
\usepackage{stfloats}

\usepackage[accepted]{icml2025}

\usepackage{amsmath}
\usepackage{amssymb}
\usepackage{mathtools}
\usepackage{amsthm}

\usepackage[capitalize,noabbrev]{cleveref}

\usepackage{multirow}

\theoremstyle{plain}

\theoremstyle{definition}

\theoremstyle{remark}

\usepackage[textsize=tiny]{todonotes}

\newcommand{\method}{Ltri-LLM}

\icmltitlerunning{\method: Streaming Long Context Inference for LLMs with Training-Free Dynamic Triangular Attention Pattern}

\begin{document}

\twocolumn[
\icmltitle{\method: Streaming Long Context Inference for LLMs with Training-Free Dynamic Triangular Attention Pattern}



\icmlsetsymbol{equal}{*}

\begin{icmlauthorlist}
\icmlauthor{Hongyin Tang}{equal,meituan}
\icmlauthor{Di Xiu}{equal,meituan,aircas}
\icmlauthor{Lanrui Wang}{meituan,iie}
\icmlauthor{Xiurui Geng}{aircas}
\icmlauthor{Jingang Wang}{meituan}
\icmlauthor{Xunliang Cai}{meituan}
\end{icmlauthorlist}

\icmlaffiliation{meituan}{Meituan Inc., Beijing, China}
\icmlaffiliation{aircas}{Aerospace Information Research Institute, Chinese Academy of Sciences, Beijing, China}
\icmlaffiliation{iie}{Institute of Information Engineering, Chinese Academy of Sciences, Beijing, China}

\icmlcorrespondingauthor{Hongyin Tang}{tanghongyin@meituan.com}

\icmlkeywords{Machine Learning, ICML}

\vskip 0.3in
]



\printAffiliationsAndNotice{\icmlEqualContribution} 

\begin{abstract}
The quadratic computational complexity of the attention mechanism in current Large Language Models (LLMs) renders inference with long contexts prohibitively expensive. 
To address this challenge, various approaches aim to retain critical portions of the context to optimally approximate Full Attention (FA) through Key-Value (KV) compression or Sparse Attention (SA), enabling the processing of virtually unlimited text lengths in a streaming manner. However, these methods struggle to achieve performance levels comparable to FA, particularly in retrieval tasks.
In this paper, our analysis of attention head patterns reveals that LLMs' attention distributions show strong local correlations, naturally reflecting a chunking mechanism for input context. We propose Ltri-LLM framework, which divides KVs into spans, stores them in an offline index, and retrieves the relevant KVs into memory for various queries. Experimental results on popular long text benchmarks show that Ltri-LLM can achieve performance close to FA while maintaining efficient, streaming-based inference.
\end{abstract}

\section{Introduction}

Recently, Large Language Models (LLMs) have made significant progress in extending context windows, as demonstrated by GPT-4 \citep{openai2024gpt4technicalreport} and LLaMA-3.1 \citep{dubey2024llama3herdmodels} with a 128K window, GLM-4-9B \citep{glm2024chatglm} with a 1M window, and Gemini-1.5-pro  \citep{geminiteam2024gemini15unlockingmultimodal} with a 10M window. 
Despite these advancements, many open-source LLMs that support extended context lengths still face substantial computational challenges during inference. These challenges stem primarily from the excessive Key-Value (KV) cache storage requirements and the quadratic computational complexity \citep{vaswani2017attention} of attention mechanisms.

KV cache compression has emerged as a promising approach to alleviate computational and memory bottlenecks during inference, with studies demonstrating its potential to improve efficiency \citep{zhang2023h2oheavyhitteroracleefficient, li2024snapkv, lee2024infinigenefficientgenerativeinference, liu2024retrievalattentionacceleratinglongcontextllm, ge2024modeltellsdiscardadaptive, adnan2024keyformerkvcachereduction, cai2024pyramidkvdynamickvcache}.
While these methods typically rely on heuristic rules to compress or discard less critical KV elements, their effectiveness is often limited by accuracy trade-offs and constrained applicability in specific contexts.

Recently, a novel approach called InfLLM has been proposed \citep{xiao2024infllm}, which introduces a block-based streaming mechanism, which can be likened to constructing a Full-Text Index while simultaneously performing Retrieval-Augmented Generation (RAG) \citep{gao2024retrievalaugmentedgenerationlargelanguage}.
Although this design aligns closely with human intuition for managing large volumes of information—storing older information and retrieving it as needed, its effectiveness is heavily influenced by the quality of text index construction and retrieval processes, underscoring the importance of robust indexing and retrieval mechanisms in achieving optimal results.

\begin{figure}[htbp]
\begin{center}
\centerline{\includegraphics[width=\columnwidth]{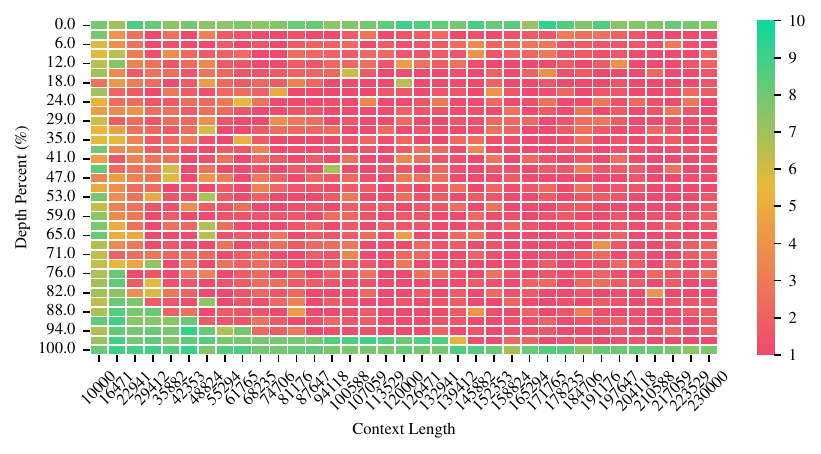}}
\caption{Average Needle-In-A-Haystack performance of InfLLM on a set of NIAH tests consisting of eight questions. More details can be found in Appendix \ref{appendix::fan}.}
\label{fig:avg_infllm_l38i262k_niah}
\end{center}
\vskip -0.3in
\end{figure}

We evaluate InfLLM on the LLAMA3-8B-Instruct-262K \footnote{https://huggingface.co/gradientai/Llama-3-8B-Instruct-262k} model on a set of Needle-In-A-Haystack(NIAH) tests consisting of eight questions. As depicted in Figure \ref{fig:avg_infllm_l38i262k_niah}, InfLLM struggles to provide correct answers. 
Our analysis revealed that the majority of InfLLM's failed cases originate from overlooking of the crucial tokens with low attention scores.

In this paper, we address this issue by preserving independent semantic spans to the greatest extent possible. Our findings reveal that the attention distributions of LLMs exhibit strong local correlations, manifesting as multiple triangular regions in the attention map. These triangular patterns naturally correspond to the model’s interpretation of semantic segmentation. Building on this observation, we employ the Non-Maximum Suppression (NMS) technique to develop a simple yet effective algorithm for identifying the boundaries of these spans. Subsequently, we dynamically generate index vectors for each span by leveraging a “voting” mechanism among its neighboring spans.

In summary, our contributions are as follows:
\begin{itemize}
    \item We investigate the reason why InfLLM struggles in NIAH test and identify that in the correct cases, the model's recall is significantly higher than the wrong cases, and the recall varies considerably across different layers. Further, we attempt to analyze the effect of mandatory evidences injection, we found it can improve the accuracy, but it still largely depends on the model’s inherent capabilities.
    \item Referring to the steps of InfLLM, we propose a novel method \textbf{\method}, which can efficiently and flexibly identifies semantic spans of the long context and store them in an offline cache, then retrieve them accurately when needed.
    \item We employ LLAMA3-8B-Instruct-262K as the base model and conduct thorough evaluation on the long context benchmarks, including NIAH, $\infty$-Bench \citep{zhang2024inftybench} and RULER \citep{hsieh2024ruler}.  Experimental results show that \textbf{\method} can achieve promising performance while maintaining equally computation cost as InfLLM.
\end{itemize}

\section{Preliminaries}
In this section, we firstly introduce the overall framework of InfLLM, including the KV offloading and retrieval process. Then, through the analysis of recall and success rate on NIAH test, we prove that the bottleneck of InfLLM lies in the performance of retrieval.

\textbf{Background.} InfLLM splits the tokens into three parts in order, i.e., the Initial tokens $\Lambda$ of length $L_{init}$, the Evicted tokens $\mathcal{E}$ of length $L_{ret}$ and the Local tokens $\mathcal{L}$ of length $L_{win}$. $\mathcal{L}$ are responsible for modeling local dependencies and the $\Lambda$ (also known as Attention Sinks) maintain the numerical stability of the attention distribution. InfLLM processes the tokens in a streaming manner. At each step, it retrieves constant number of relevant blocks as $\mathcal{E}$ from the context memory using the similarity score and apply attention to the concatenation of $\Lambda, \mathcal{E}, \mathcal{L}$. When the length of the input tokens exceeds $L_{win}$, it stores tokens from the beginning of $\mathcal{L}$ to context memory, picking tokens with high attention scores as the index vectors.

\begin{figure}[htbp]
\vskip -0.15in
    \subfloat[Average on 11 successful samples with 26 decoding steps.]{%
        \includegraphics[width=.48\linewidth]{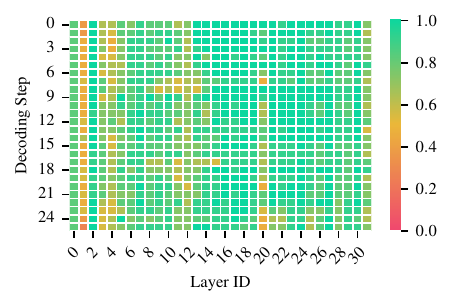}%
        \label{fig:success_needle7_steps26_samples11}%
    }\hfill
    \subfloat[Average on 212 failed samples with 19 decoding steps.]{%
        \includegraphics[width=.48\linewidth]{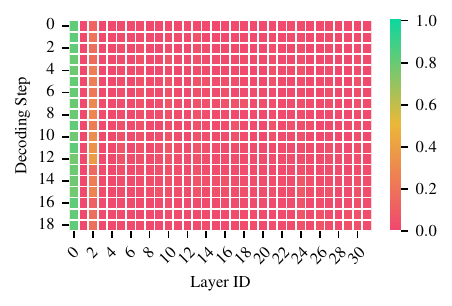}%
        \label{fig:failure_needle7_steps19_samples212}%
    }
    \caption{The average needle recall across layers and decoding steps of successful and failed NIAH cases.}
    \label{fig:niah_dist_evidence_inclusion}
\end{figure}

\textbf{Response Quality Hinges on Recall.} We construct a group of NIAH tests with eight bilingual questions. More details are shown in Appendix \ref{appendix::fan}. We carefully record whether the needle is incorporated in the retrieved blocks or not at each decoding step. The model's responses are scored by GPT-4  \citep{openai2024gpt4technicalreport}, only the responses with the highest score are considered as successful ones.
We plot the average recall across different decoder layers and decoding steps. Figure \ref{fig:niah_dist_evidence_inclusion} demonstrates the needle recall for both successful and failed instances. Obviously, there is a discernible trend shows that the successful cases have a much higher recall comparing to the failed cases.

\textbf{Mandatory Evidence Injection Transforms Unreasonable Responses.} The observed relationship raises the question of whether increasing the recall could transform failed cases into successful ones. We validate this hypothesis by mandatorily injecting the blocks containing the needle into $\mathcal{E}$ on the failed cases. The results of conversion are shown in Table \ref{tab:fan_custom_niah}. After the mandatory injection, nearly all English NIAH cases are successfully converted, whereas the lower conversion rate of Chinese NIAH cases is probably due to the tested LLAMA-3 model being predominantly trained on English corpus. This phenomenon confirms that the bottleneck of InfLLM lies in the retrieval recall.

\begin{table}[htbp]
\caption{Conversion results of mandatory evidence injection on NIAH tests with different genres.}
\label{tab:fan_custom_niah}
\vskip -0.15in
\begin{center}
\resizebox{\columnwidth}{!}{
\begin{NiceTabular}{cccccc}
\CodeBefore
  \rowcolors{2}{white}{lightgray!20}
\Body
\toprule
\textbf{Task ID (Zh)}  & \textbf{\#Failure} & \textbf{\#Converted} & \textbf{Task ID (En)} & \textbf{\#Failure} & \textbf{\#Converted} \\
\midrule
    1    & 477  & 51  & 5  & 894 & 773 \\
\midrule
    2    & 682  & 638 & 6  & 775 & 773 \\
\midrule
    3    & 1021 & 2   & 7  & 212 & 210 \\
\midrule
    4    & 211  & 0   & 8  & 826 & 702 \\
\bottomrule
\end{NiceTabular}
}
\end{center}
\vskip -0.2in
\end{table}

\section{Method}
\subsection{Local Triangle-shaped Aggregated Attention Pattern}

\begin{figure}[H]
\vskip -0.15in
    \subfloat[\textit{Layer 8}]{%
        \includegraphics[width=.48\linewidth]{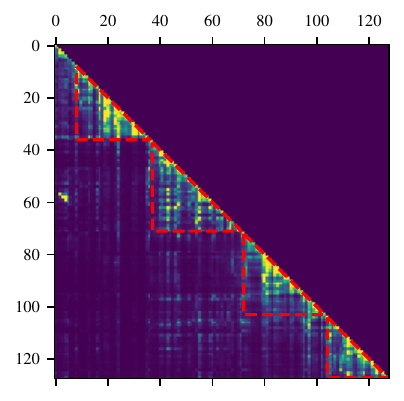}%
        \label{fig:triangular_attention_layer_8}%
    }\hfill
    \subfloat[\textit{Layer 14}]{%
        \includegraphics[width=.48\linewidth]{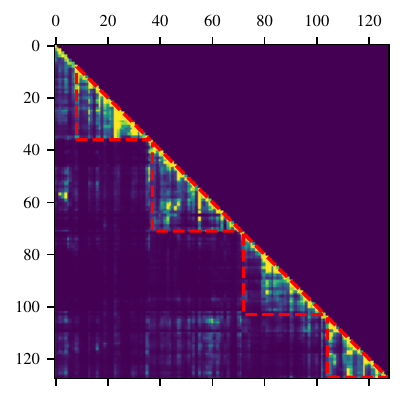}%
        \label{fig:triangular_attention_layer_14}%
    }\\
    \subfloat[\textit{Layer 20}]{%
        \includegraphics[width=.48\linewidth]{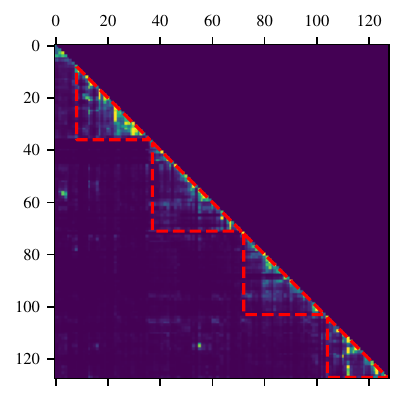}%
        \label{fig:triangular_attention_layer_20}%
    }\hfill
    \subfloat[\textit{Layer 26}]{%
        \includegraphics[width=.48\linewidth]{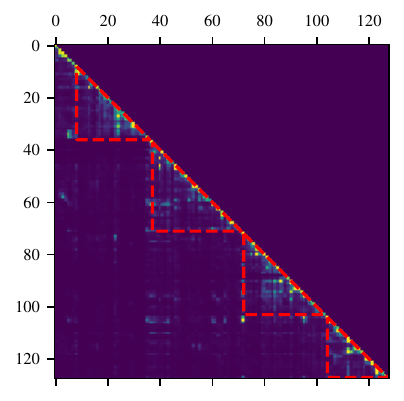}%
        \label{fig:triangular_attention_layer_26}%
    }\\
    \caption{Triangular attention regions indicate semantic segments.}
    \label{fig:triangle_attention_pattern}
\vskip -0.2in
\end{figure}


Figure \ref{fig:triangle_attention_pattern} illustrates the attention map across different layers of the model. Tokens within the triangular regions exhibit higher mutual attention scores compared to those outside these regions. This phenomenon, commonly referred to as localized attention or block-sparse attention, has been highlighted in recent studies such as PyramidKV \citep{cai2024pyramidkvdynamickvcache} and MInference \citep{jiang2024minference}. Leveraging this property, we construct indexes for distinct semantic spans, aiming to retain more semantic information while minimizing the associated memory footprint.

\subsection{Semantic Span Division and NMS Acceleration}\label{sec:3.2}
\begin{figure}[htb]
\begin{center}
\centerline{\includegraphics[width=\columnwidth]{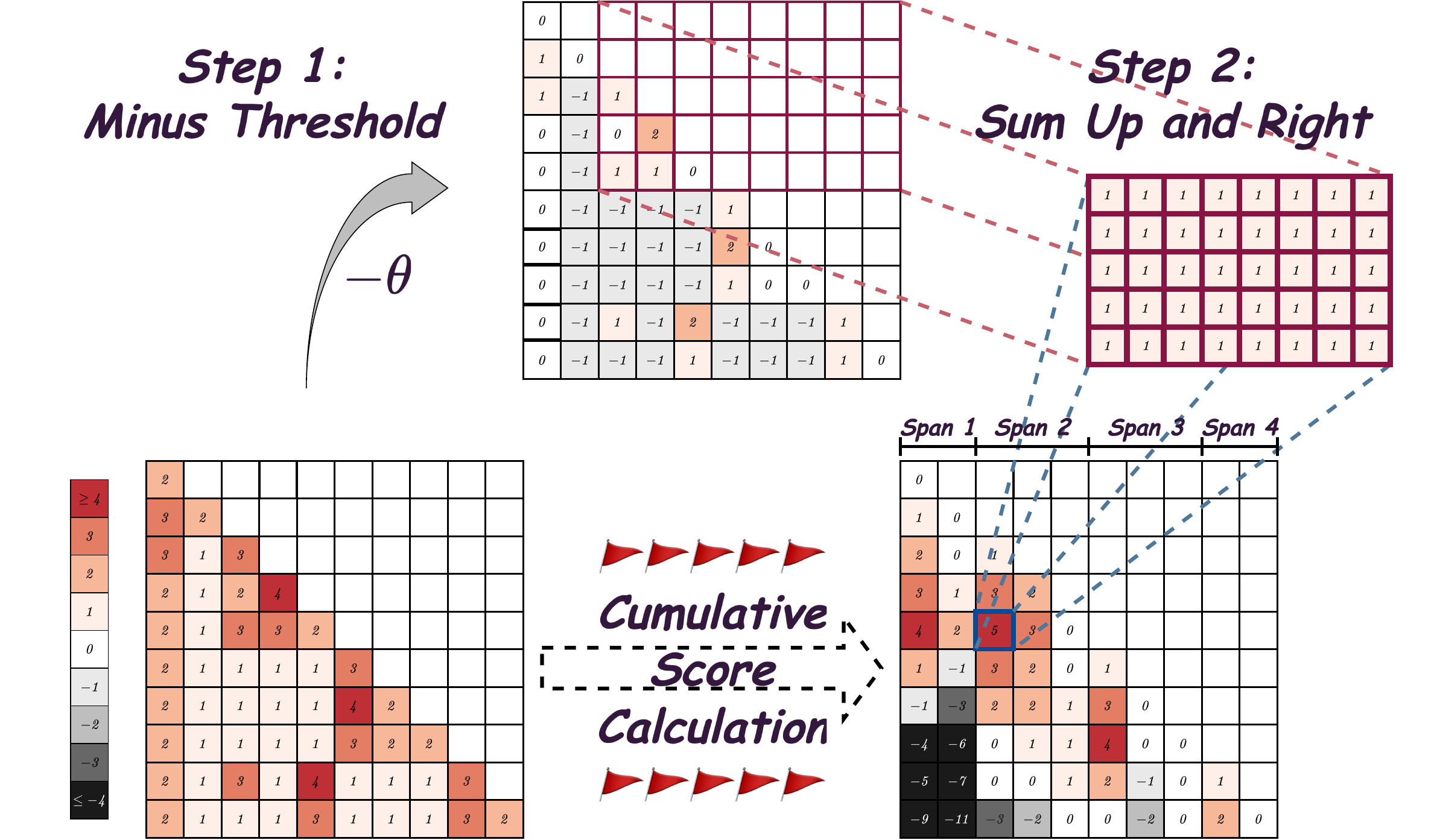}}
\caption{Cumulative scores calculation for each span. First, minus a threshold $\theta$. Second, sum the upper and right elements.}
\label{fig:calc_cumulative}
\end{center}
\vskip -0.3in
\end{figure}

We develop an algorithm that can fast determines the boundaries of these triangle regions. The formulation of the algorithm is given as follows. Given Query $Q_{H \times N\times d}$ and Key $K_{H\times N\times d}$, where $H$ is the number of heads, $N$ is the length of tokens, and $d$ is the dimension of each head, the attention map of Multi-Head Attention (MHA) of a single head $h\in[1,H]$, $A_{h}$ is defined as:
 \begin{equation}
    A_h^{*} = \texttt{softmax}(\frac{Q_h K_h^T}{\sqrt{d}})
 \end{equation}

To prevent the earlier tokens from attending to the subsequent ones in auto-regressive LLMs,  a lower triangular matrix mask is applied to the attention map. The mask matrix $P_{N\times N}$ is defined as:
\begin{equation}
    P_{N\times N}=\{p_{ij}|{1\leq i\leq N,1 \leq j \leq N}\}= 
\begin{cases} 
1, & \text{if } i \geq j \\
0, & \text{if } i < j
\end{cases}
\end{equation}
Then the final attention map $A_{h}$is:
\begin{equation}
    A_{h} = A_h^{*} \circ P = \begin{cases} 
a_{hij}, & \text{if } i \geq j \\
0, & \text{if } i < j
\end{cases}
\end{equation}

Next, we sum the attention scores over heads, and define the Triangle Attention (TA) score of span between $x$ and $y$ as:
\begin{equation}
    S^{*}_{xy}(0\leq x \leq y \leq N) = \sum_{h=1}^H\sum_{i=x}^y\sum_{j=x}^y A_{hij} =  \sum_{h=1}^H\sum_{i=0}^y\sum_{j=x}^N A_{hij}
    \label{eq:4}
\end{equation}

Since the upper diagonal elements of  $A_h$  are zero, we can apply an accumulation summation operator, such as \texttt{cumsum}, as shown in the second equation of Eq.\ref{eq:4}. Notably, as the elements of the attention map must be positive (as per Eq.\ref{eq:4}), the largest TA score is  $S_{0N}$ , located at the bottom-left corner of the attention map. The triangular regions stand out due to the significant score differences compared to their surrounding areas. Inspired by the concept of Baseline Reward in the REINFORCE algorithm \citep{williams1992simple}, we introduce a threshold $\theta$ for the attention map. By subtracting $\theta$, scores below this threshold become negative, allowing us to effectively filter out unwanted TA scores and retain only the relevant ones:

\begin{equation}
    S_{xy}(0\leq x \leq y \leq N) = \sum_{h=1}^H\sum_{i=0}^y\sum_{j=x}^N (A_{hij} - P*\theta)
    \label{eq:sxy}
\end{equation}

From Eq.\ref{eq:sxy}, we can get the TA score for every span of text. Unfortunately, the spans with high TA scores are highly overlapped (e.g., a subset of a salient triangle region is also likely to be another salient triangle region).

To address this issue, the Non-Maximum Suppression (NMS) algorithm \citep{ren2017faster}, originally designed to eliminate redundant boxes in object detection models, is employed to remove overlapping spans with low scores. As NMS is widely utilized in various efficient on-device models, we adopted the off-the-shelf implementation as part of our method. Further details of the algorithm are provided in Appendix \ref{appendix::span_selection_alg}.

\begin{figure}[htb]
\begin{center}
\centerline{\includegraphics[width=\columnwidth]{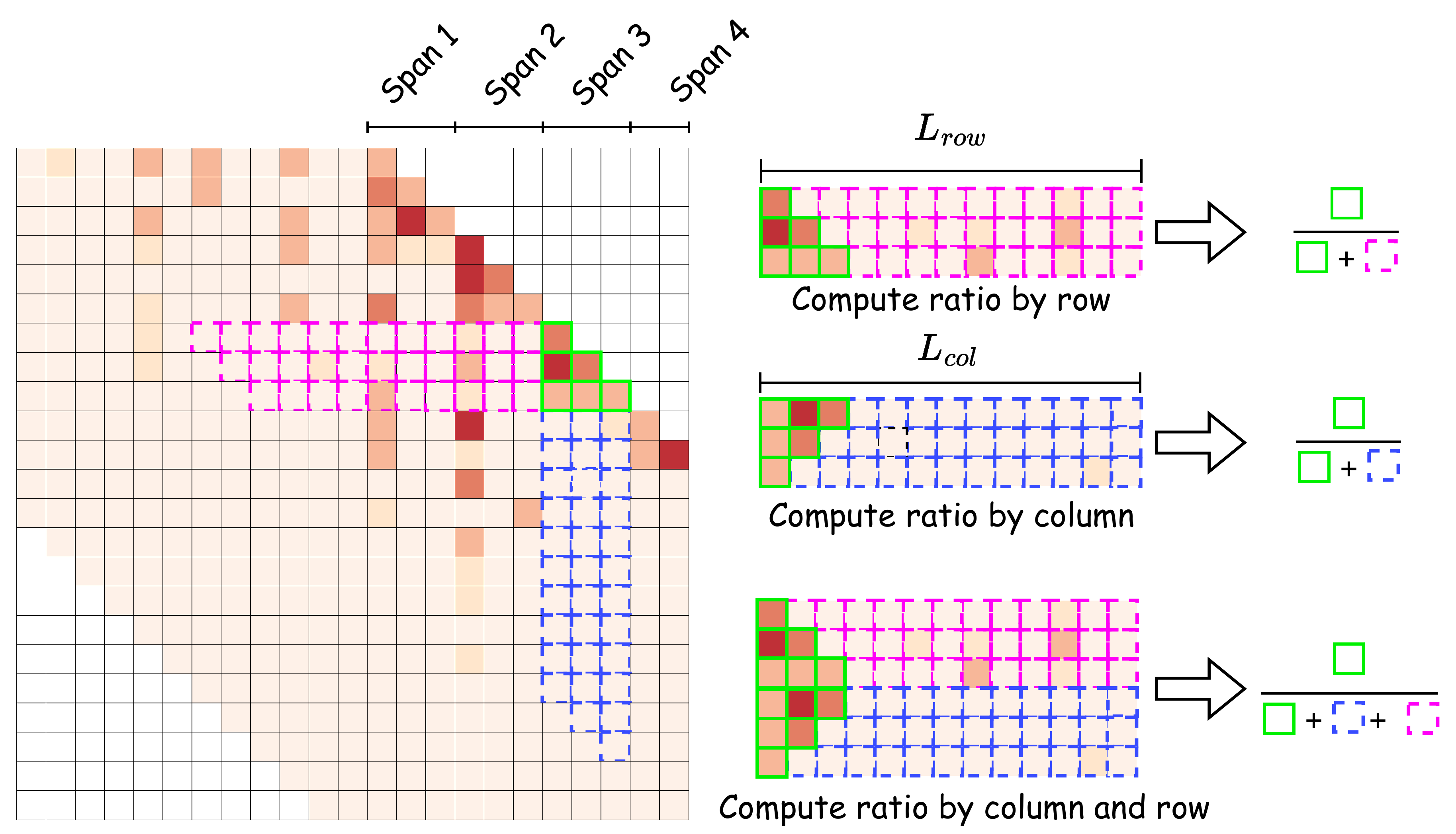}}
\caption{Three metrics gauge the confidence of the voters, i.e., the ratio between the span and its row neighbors, its column neighbors and its row+column neighbors.}
\label{fig:ratio}
\end{center}
\vskip -0.3in
\end{figure}
\subsection{Span Index Vector}\label{sec:3.3}
In this section, we will discuss the static and dynamic methods of how to generate the index vectors of the evicted tokens $\mathcal{E}$. The static methods generate constant number of index vectors for each span whereas the dynamic methods generate a dynamic number of index vectors based on the confidence score of each span. 

\subsubsection{Static Span Index Vectors}
InfLLM adopts a static method which calculates the cumulative attention score for each token. Formally, letting $v$ be the accumulated attention score and $L_{win}$ be the sliding window size, the accumulation score of token $x$ is,
\begin{equation}
    v_x = \sum_{h=1}^H \sum_{i=1}^{L_{win}}A_{hix}
\end{equation}
InfLLM adopts the Keys of top $C$ tokens with the largest $v$ as the index vectors of a fixed-size block. We use this method to generate $C/N_s$ index vectors for each span, where $N_s$ is the number of spans for each block. We also tried to use the mean pooling of the Keys within each span as the index vector. Unfortunately, we found that the static methods performed poorly in the early experiments. 

\subsubsection{Dynamic Span Index Vectors}\label{sec:3.3.2}

The static method treats the accumulated attention scores acquired by a token as a form of voting, suggesting that tokens with the highest votes can effectively represent a span.

However, in certain cases, the voters may lack sufficient knowledge about the candidates, making it difficult to identify truly outstanding ones. When voter confidence is low, retaining more index vectors for a span becomes crucial to ensure accurate retrieval. To address this, three metrics are introduced to evaluate voter confidence, enabling the allocation of fewer index vectors to spans with higher confidence and more to those with lower confidence.


The three metrics are based on the ratio of the TA score and its neighbors. As illustrated in Figure \ref{fig:ratio}, consider a block divided into four spans (\texttt{Span1} to \texttt{Span4}), with \texttt{Span3} as an example. The ratio of the area between the span and its neighbors is computed. 
As shown in Figure \ref{fig:ratio}, assuming a block is divided into four spans (i.e., \texttt{Span1} to \texttt{Span4}) and taking \texttt{Span3} as an example, we compute the ratio of the area between the span and its neighbors. 
Three specific ratios are defined to capture the relationships of \texttt{Span3} with its row neighbors, column neighbors, and both. 
A high ratio suggests that the neighbors have limited knowledge about this span, indicating insufficient voting confidence, necessitating more index vectors for this span. Conversely, a low ratio implies sufficient confidence, allowing us to represent the span with fewer index vectors. 

\begin{figure}[htbp]
\vskip -0.15in
\begin{center}
\centerline{\includegraphics[width=\columnwidth]{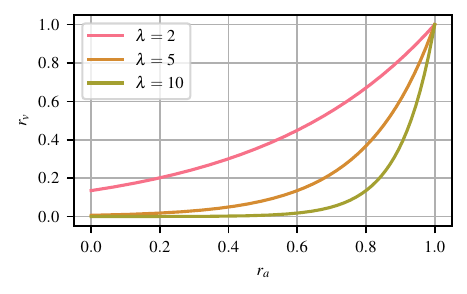}}
\caption{The value of $r_v$ for different $\lambda$}
\label{fig:lambda_ratio}
\end{center}
\vskip -0.3in
\end{figure}

Next, we develop a specific function to establish a correlation between the level of confidence and the number of index vectors to be retained. Specifically, we use a function where the input variable is the ratio of the span's area to its neighbors' area, denoted as $r_a$, and the output variable is the ratio of index vectors to be retained, denoted as $r_v$. Formally, $r_v$ is defined as follows:
\begin{equation}
    r_v = \frac{1}{e^{\lambda(1-r_a)}}
    \label{eq:rv}
\end{equation}

where $\lambda$ is a hyper-parameter controlling the increasing speed of the function. As shown in Figure \ref{fig:lambda_ratio}, a larger $\lambda$ imposes a stronger restriction on the number of index vectors. The number of the index vectors is calculated by the length of span times the $r_v$, i.e., $L_{span}\times r_v$. In practice, we often set a minimal and maximal number of index vector to restrict the number of index vectors stored in GPU Memory.

\begin{figure*}[t]
\begin{center}
\centerline{\includegraphics[width=1.8\columnwidth]{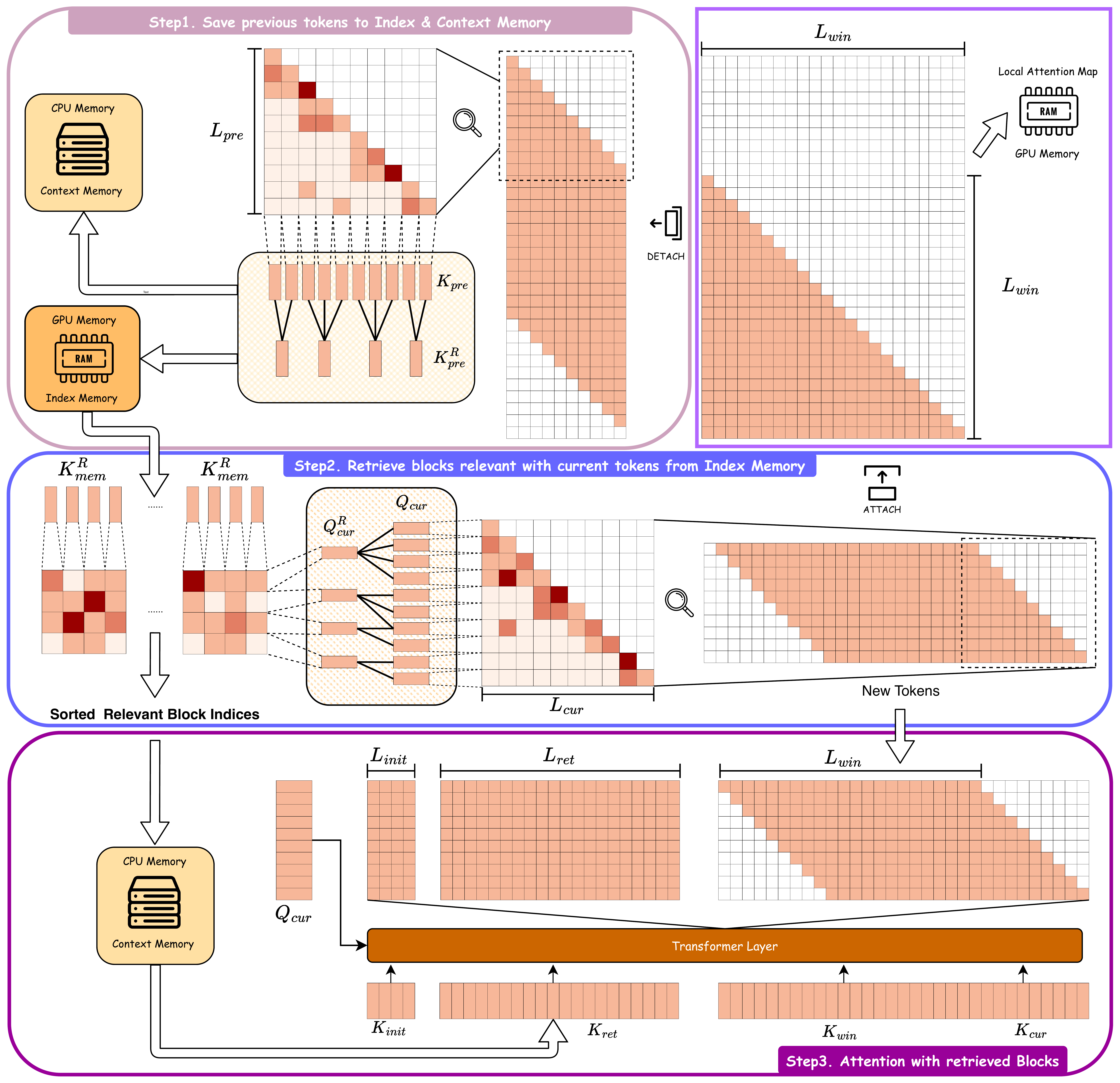}}
\caption{Flowchart of Ltri-LLM framework.}
\label{fig:flowchart}
\end{center}
\vskip -0.3in
\end{figure*}

In LLMs, the distribution of area ratios across different layers can vary significantly, making it challenging to control the number of index vectors within a reasonable range using a fixed $\lambda$. To determine an appropriate $\lambda$ for each layer, we first collect the values of $r_a$ under different contexts and calculate the probability of $r_a$ across various ranges. Using these probabilities, we can calculate the expected number of index vectors for any given $\lambda$ through Eq. \ref{eq:rv}. Consequently, for any specified compression ratio (i.e., the ratio between the number of tokens and the number of index vectors), we can find the smallest $\lambda$ that meets this compression ratio for each layer by consulting a lookup table. This approach retains more index vectors while ensuring the desired compression ratio, thereby improving recall.

\subsection{Overall Framework}

In this section, we will introduce the Ltri-LLM framework step-by-step. Ltri-LLM mainly consists of three steps as shown in Figure \ref{fig:flowchart}. In Ltri-LLM, we process the input context in a streaming manner with a fixed window size. The streaming-type fixed-size Local Attention Map is always stored in the GPU Memory as shown in the upper right side of Figure \ref{fig:flowchart}. As the process goes on, we save the previous tokens to the Context Memory and then retrieve them as needed. Suppose we need to process a new block of context whose length is $L_{cur}$. 

In the first step, we detach the left part of the attention map which is corresponding to the previous $L_{pre}$ tokens in the GPU Memory. Next, we apply the method introduced in Section \ref{sec:3.2} to divide the $L_{pre}$ into $C$ semantic individual spans. Then, we use the approach from Section \ref{sec:3.3} to generate the index vectors $K^R_{pre}$ of these spans. We store the Keys of the previous tokens $K_{pre}$ to the CPU Memory and $K^R_{pre}$ to the GPU memory. 

In the second step, we divide the tokens of the upcoming block using the same method as in the first step, generating in the index vectors for the Queries, $Q^R_{cur}$. Then, we retrieve relevant tokens using $Q^R_{cur}$ by calculating the similarity score with the index vectors stored in the Index Memory, $K^R_{mem}$. Specifically, assuming the $i$-th span's $K_{mem}^R$ stored in the Index Memory consists of $M$ index vectors, and the $j$-th span's $Q^R_{pre}$ consists of $N$ index vectors, the similarity score between the $i$-th $K_{mem}^R$ and the $j$-th $Q^R_{pre}$, $\text{sim}_{ij}$, is computed as follows:

\begin{equation}
    \text{sim}_{i,j} = \sum_{m=1}^{M}{\sum_{n=1}^{N}{Q^R_{pre,j}[m]\cdot K^R_{mem,i}[n]}}
\end{equation}

Next, we retrieve the relevant blocks based on the similarity scores. Specifically, considering that each block consists of $C$ spans and there are $B$ blocks stored in the Index Memory, we sort the blocks by the maximum similarity of the $C$ spans. We then select the top K indices as the retrieval results, denoted as $R_{ret}$.
\begin{equation}
    R_{ret} = \text{Top}_{b\in(0,B]} (\max\{\text{sim}_{ij}|i\in[b_l,b_r],j\in(0,C]\}), K)
\end{equation}
where $[b_l,b_r]$ indicates the indices of spans corresponding to block $b$. 

In the third step, we perform attention between $Q_{cur}$ and the retrieved results. Specifically, we begin by concatenating the Keys of the initial $L_{init}$ tokens, the retrieved $L_{ret}$ tokens, the nearest $L_{win}$ tokens, and the current $L_{cur}$ tokens. Then, we use $Q_{cur}$ to attend to these Keys. The outputs of this attention process are passed to the next layer for further computation, and the attention map of $Q_{cur} \times [K_{win}, K_{cur}]$ is appended to the fixed-size local attention map stored in the GPU memory.

\subsection{Accurate and Persistent Collaborative Evidence Retrieval}\label{sec:3.5}
In addition to the Triangle pattern, we also identified three components that can improve the retrieval accuracy, including Persistent Mechanism, Retrieval Heads, and Collaborative Voting.

\textbf{Persistent Mechanism.} 
As discussed in previous sections, the performance of Ltri-LLM heavily relies on the recall of the current block. However, during the decoding stage, the block becomes a single token, which can significantly amplify the variation in results. To mitigate this variation, we choose to retain the retrieved blocks from the last chunk of the prefilling stage throughout the decoding stage.

\textbf{Retrieval Heads.} 
\citet{wu2024retrievalheadmechanisticallyexplains} discovered that only a small subset of attention heads significantly contributes to retrieval ability, and these are referred to as retrieval heads. To leverage these heads, we first identify them using the method outlined by \citet{wu2024retrievalheadmechanisticallyexplains}, labeling attention heads with scores above 0.1 as retrieval heads. In our approach, when a layer contains multiple retrieval heads, we retain the one with the highest score. The selected retrieval heads and their corresponding scores are detailed in Appendix \ref{appendix::rh}.

\textbf{Collaborative Voting.} During the forward process, each layer has its own context manager and independently determines the location of evidence. When the $\mathcal{K}$-th layer needs to identify evidence, it can use the results from all preceding layers to make a comprehensive judgment. In practice, we retain the results from layers that have at least one retrieval head along with their corresponding retrieval head scores. For a thorough evidence location judgment, we suggest using a voting mechanism where the retrieval head scores act as voting weights. 

\section{Experiments}\label{experiments}
\textbf{Implementation Details.} Ltri-LLM is consistent with InfLLM on the primary hyperparameters. Specifically, the number of initial tokens, local window size, memory unit size, GPU cache size, encoding chunk size, and score decay coefficient are set at 128, 4096, 128, 32, 512, and 0.1 respectively. Unless otherwise stated, in Section \ref{sec:3.3.2}, we employ the row ratio calculation method and designate $\lambda$ as 3. The $\lambda$ value for the first four layers is assigned as 20. Furthermore, we ensure the last encoding chunk size of the prefilling stage can accomodate all query tokens without beging excessively large. While the attention map threshold $\theta$, the Intersection over Union (IoU) threshold $\varphi$ for the Non-Maximum Suppression (NMS) operator from torchvision and the maximum number of retained index vectors are detailed in Appendix \ref{appendix::configuration}.

\textbf{Dataset \& Evaluation Metrics.} We evaluate our method by three widely used benchmarks, i.e., Neelde-In-A-Haystack \citep{niah}, $\infty$-Bench \citep{zhang2024inftybench} and RULER \citep{hsieh2024ruler}. Needle-In-A-Haystack is designed to test in-context retrieval ability of long context LLMs. $\infty$-Bench comprises synthetic and realistic tasks spanning diverse domains to evaluate the comprehensive modeling capabilities for long context, presented in both English and Chinese. RULER expands upon the vanilla NIAH test to encompass variations with diverse types and quantities of needles. It also introduces new task categories including multi-hop tracing and aggregation to test behaviors beyond searching from context.

\begin{table*}[ht]
\vskip -0.1in
\caption{Performance of different methods on $\infty$-Bench. The symbol $\ddag$ indicates workaround results. Zh.QA tasks have instances with extremely long contexts, leading to CUDA out of memory (OOM) problems for InfLLM and Ltri-LLM. To avoid these OOM issues, while keeping the same number of instances for comparison with other methods, we uniformly truncate contexts to 800K when inferring Zh.QA with InfLLM and Ltri-LLM, albeit at the expense of performance degradation. The performance of symbol $\dag$ is cited from \citep{jiang2024minference}. The highest scores of non-FA and Streaming methods are highlighted in bold and underlined, respectively. }

\vskip 0.05in
\label{tab:main_results_infinitebench}
\begin{center}
\begin{small}
\resizebox{\textwidth}{!}{
\begin{NiceTabular}{lccccccccccccc}
\CodeBefore
  \rowcolors{2}{white}{lightgray!20}
\Body
\toprule
Methods & Streaming & En.Sum & Zh.QA & En.QA & En.MC & En.Dia & Co.De & Ma.Fd & Re.PK & Re.Num & Re.KV & Avg. w/o KV & Avg. \\
\midrule
LLAMA3-8B-Instruct-262K $^\dag$        & \ding{55} & 20.2 & 12.9 & 12.4 & 67.3 & 6.0 & 22.1 & 26.6 & 100.0 & 100.0 & 14.4 & 40.8 & 38.2 \\
MInference        & \ding{55} & \textbf{20.8} & \textbf{10.6} & 12.5 & \textbf{65.9} & \textbf{7.5} & 22.3 & \textbf{32.3} & \textbf{100.0} & \textbf{100.0} & 12.8 & \textbf{41.3} & 38.5 \\
LM-Infinite       & \ding{51} & 14.0 & 7.8 & 14.1 & 39.3 & 3.5 & 42.9 & 17.1 & 6.8 & 6.8 & 2.4 & 16.9 & 15.5 \\
StreamingLLM      & \ding{51} & 14.9 & 8.0 & 14.1 & 38.9 & 4.0 & 42.6 & 16.7 & 6.8 & 6.8 & 2.4 & 17.0 & 15.5 \\
InfLLM            & \ding{51} & \underline{16.3} & 8.5$^\ddag$ & 19.5 & 38.0 & \underline{6.0} & \underline{\textbf{44.7}} & 23.7 & \underline{\textbf{100.0}} & \underline{\textbf{100.0}} & 30.2 & 39.6 & 38.7 \\
\textbf{Ltri-LLM} & \ding{51} & 16.1 & \underline{9.6}$^\ddag$ & \underline{\textbf{21.2}} & \underline{44.5} & \underline{6.0} & 41.6 & \underline{26.3} & \underline{\textbf{100.0}} & \underline{\textbf{100.0}}  & \underline{\textbf{64.0}} & \underline{40.6} & \underline{\textbf{42.9}} \\
\bottomrule
\end{NiceTabular}
}
\end{small}
\end{center}
\vskip -0.2in
\end{table*}

\textbf{Baselines.} We compare our method with four training-free baselines, including three streaming  methods: LM-Infinite \citep{han2023lm}, StreamingLLM \citep{xiao2023streamingllm} and InfLLM \citep{xiao2024infllm} as well as a Non-streaming state-of-the-art long context accelerating method, MInference \citep{jiang2024minference}, which retains the full attention between the last few tokens and the whole context. In LM-Infinite and StreamingLLM, we specify the the number of initial tokens and local tokens as 128 and 6144, respectively. For InfLLM, the number of initial tokens, local tokens, block size, the number of representative tokens and reserved blocks are set to 128, 4096, 128, 4, and 16, respectively. These configurations facilitate a fair comparison in terms of the number of tokens involved in the attention calculation. For MInference, we use the default configuration recipe for LLAMA3-8B-Instruct-262k.

\begin{figure}[htbp]
    \subfloat[InfLLM, w/o \textit{restrict}]{%
        \includegraphics[width=.48\linewidth]{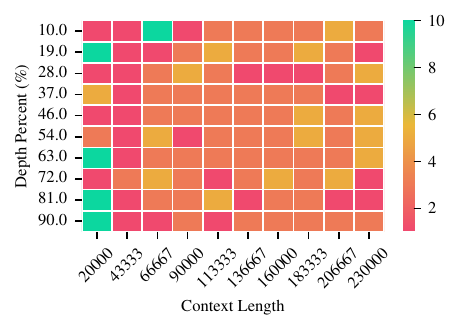}%
        \label{fig:infllm_disable_restrict_needle_within_block}%
    }\hfill
    \subfloat[InfLLM, w/ \textit{restrict}]{%
        \includegraphics[width=.48\linewidth]{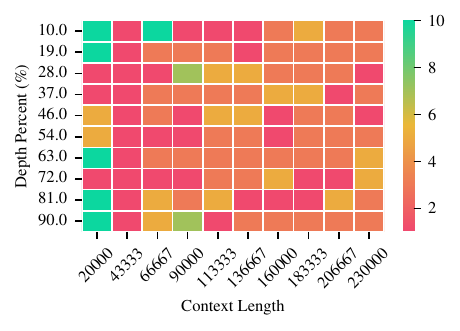}%
        \label{fig:infllm_enable_restrict_needle_within_block}%
    }\\
    \subfloat[Ltri-LLM, w/o \textit{restrict}]{%
        \includegraphics[width=.48\linewidth]{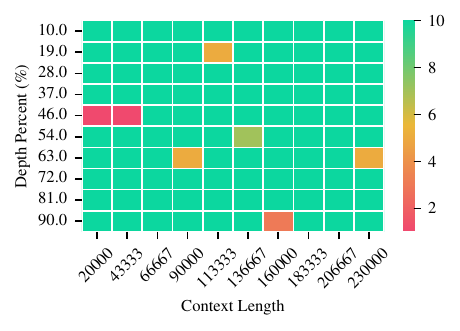}%
        \label{fig:ltrillm_disable_restrict_needle_within_block}%
    }\hfill
    \subfloat[Ltri-LLM, w/ \textit{restrict}]{%
        \includegraphics[width=.48\linewidth]{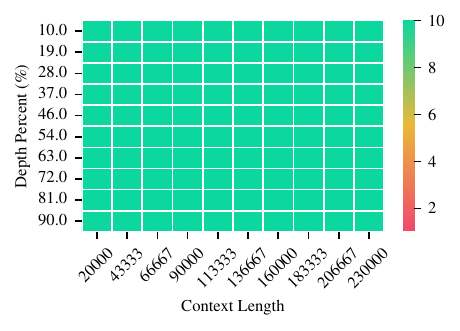}%
        \label{fig:ltrillm_enable_restrict_needle_within_block}%
    }
    \caption{Performance on Needle-In-A-Haystack Benchmark.}
    \label{fig:comparison_niah}
\end{figure}

\textbf{Needle-In-A-Haystack.}
We set the test length of NIAH to span between 20K and 230K, with the needle's position varying from 10\% to 90\%. Each NIAH experiment is conducted across 100 groups. In both InfLLM and Ltri-LLM, the needles might be split into multiple blocks, possibly impacting performance. To evaluate this effect, we impose a restriction that the needle should locate within one single block when creating the corpus, and compare the performance before and after applying this restriction. Figure \ref{fig:comparison_niah} demonstrates that Ltri-LLM consistently outperforms InfLLM under both conditions and exhibits strong robustness even when the restriction is removed.

\textbf{$\infty$-Bench.} Table \ref{tab:main_results_infinitebench} shows the results of $\infty$-Bench. Ltri-LLM performs exceptionally well across retrieval tasks like Retr.PassKey, Retr.Number and Retr.KV. Ltri-LLM has outperformed all baseline models on the Retr.KV task, even significantly outshining the vanilla model of LLAMA3-8B-Instruct-262K. This suggests that Ltri-LLM has managed to retain crucial information while filtering out noise, thereby minimizing any distractions to the model. For other more realistic tasks such as En.QA and Zh.QA, Ltri-LLM achieves the best among all models and the best among streaming models, respectively. Overall, Ltri-LLM has outshone all other models by achieving the highest total average score in all 10 tasks, achieving the best among all models in 4 tasks, and the best among streaming models in 8 tasks. Even without the best-performing Retr.KV, Ltri-LLM can still achieve performance close to that of vanilla LLAMA3-8B-Instruct-262K and MInference. Given that MInference needs to retain the full attention of the last part of the input tokens and all previous input tokens, Ltri-LLM's streaming manner offers greater potential for efficiency. 

\begin{table*}[htb]
\centering
\caption{Performance of different methods on RULER evaluated at lengths from 4K to 128K. Results of symbol $\dag$ are cited from \citep{jiang2024minference}. The highest scores of non-FA and Streaming methods are highlighted in bold and underlined, respectively.}
\resizebox{0.6\textwidth}{!}{
\begin{NiceTabular}{lcccccccc}
\CodeBefore
  \rowcolors{2}{white}{lightgray!20}
\Body
\toprule
Methods & Streaming & 4K & 8K & 16K & 32K & 64K & 128K & Avg. \\
\midrule
LLAMA3-8B-Instruct-262K$^\dag$ & \ding{55} & 97.2 & 91.8 & 87.3 & 80.8 & 77.4 & 72.2 & 84.4 \\
MInference & \ding{55} & \textbf{92.8} & \textbf{87.2} & \textbf{87.3} & \textbf{82.3} & \textbf{82.8} & \textbf{77.1} & \textbf{84.9} \\
LM-Infinite & \ding{51} & 92.5 & 65.9 & 44.2 & 23.2 & 16.8 & 10.7 & 42.2 \\
StreamingLLM & \ding{51} & 92.5 & 65.6 & 44.0 & 22.9 & 16.8 & 11.5 & 42.2 \\
InfLLM & \ding{51} & 92.5 & 83.6 & 59.0 & 36.5 & 31.2 & 26.7 & 54.9 \\
\textbf{Ltri-LLM} & \ding{51} & \textbf{92.8} & \underline{83.7} & \underline{76.6} & \underline{72.3} & \underline{67.6} & \underline{66.7} & \underline{76.6} \\
\bottomrule
\end{NiceTabular}
}
\label{tab:ruler_results}
\end{table*}

\begin{figure}[htbp]
\begin{center}
\centerline{\includegraphics[width=\columnwidth]{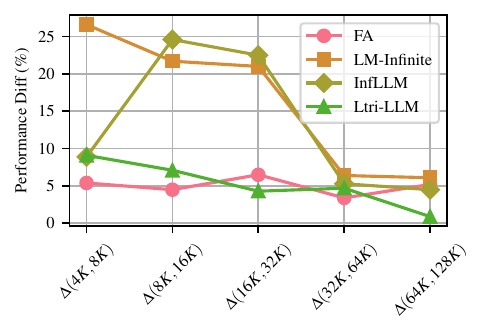}}
\caption{Performance deterioration on RULER benchmark with increasing context length. The performance difference between 4K and 8K, denoted as $\Delta(4K, 8K)$, is calculated by subtracting the performance of 8K from the 4K's. Similar to Full Attention (FA), Ltri-LLM exhibits mild performance decrease, unlike the severe metric drop observed in other streaming methods.}
\label{fig:ruler_diff_trend}
\end{center}
\end{figure}

\textbf{RULER.} Models are evaluated with context sizes ranging from 4K to 128K, and the average metric of all tasks is reported. Compared to other streaming baselines, Ltri-LLm has achieved a significant lead at ranging from 4K to 128K context lengths. 
Another noteworthy phenomenon is that the performance of almost all methods will decrease as the evaluation length increases. As shown in the Figure \ref{fig:ruler_diff_trend}, we draw the delta value between adjacent lengths in RULER for different models. It's particularly noticeable how the streaming baselines take a steeper dive, while the performance drop for Ltri-LLM and vanilla LLAMA-3-8B-Instruct-262K remains fairly consistent. This reflects that previous streaming baselines have difficulty in scaling up to longer texts, and Ltri-LLM has truly overcome this problem. 
Although Ltri-LLM has a clear advantage over streaming baslines, there is still a gap with MInference. As shown in the Figure \ref{fig:detailed_comparison_with_MInference_on_RULER}, we compared the scores of Ltri-LLM and MInference on different tasks in RULER. The complete scores on every length can be viewed in the Appendix \ref{appendix::detailed_comparison_with_MInference_on_RULER}. Ltri-LLM basically tied with MInference in the single NIAH test, but there was a noticeable gap in the more difficult multi-key NIAH test and variable tracking tasks. 
We suspect that this shortcoming might be due to the streaming manner of the Ltri-LLM. In other words, Ltri-LLM is unaware of the final question during the phase of streaming context encoding. Although Ltri-LLM has made efforts to enhance its retrieval capabilities through TA, it still falls short when put under rigorous tests like the RULER, which demands high retrieval prowess. 

\section{Ablation Studies \& Analysis}\label{sec::ablation_and_analysis}
\textbf{Ablation Study.} We conduct ablation studies on the NIAH test to evaluate the three components proposed in \ref{sec:3.5}. During these studies, the ratio mode $\Omega$, prefilling last chunk size $\mathcal{Q}$ and $\lambda$ for dynamic span index vectors are assigned to row, 32 and 3, respectively. Our main observations are: (1) The retrieval heads are crucial for performance. Disabling retrieval heads and using average results from all attention heads for retrieval hinders performance, suggesting that the other attention heads except retrieval heads, act as noise disturbances. (2) The persistent mechanism and the voting mechanism contribute positively. Notably, when performing ablation study, we enable the needle synthetic location \textit{restrict} as mentioned in Figure \ref{fig:comparison_niah}.

\begin{figure}[htbp]
    \subfloat[$\widetilde{\mathcal{RH}},\widetilde{\mathcal{P}},\widetilde{\mathcal{V}}$]{%
        \includegraphics[width=.24\linewidth]{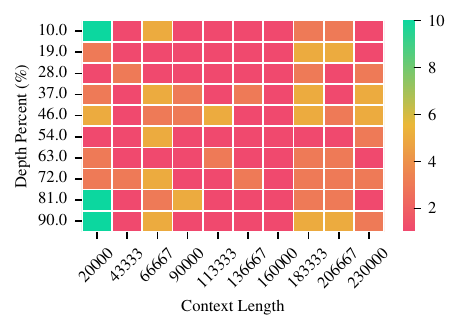}%
        \label{fig:ltrillm_RH_none_Persistent_none_Voting_none_restrict_needle_within_block}%
    }\hfill
    \subfloat[$\widetilde{\mathcal{RH}},\widetilde{\mathcal{P}},\mathcal{V}$]{%
        \includegraphics[width=.24\linewidth]{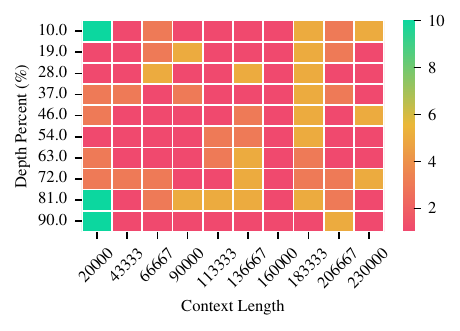}%
        \label{fig:ltrillm_RH_none_Persistent_none_Voting_restrict_needle_within_block}%
    }\hfill
    \subfloat[$\widetilde{\mathcal{RH}},\mathcal{P},\widetilde{\mathcal{V}}$]{%
        \includegraphics[width=.24\linewidth]{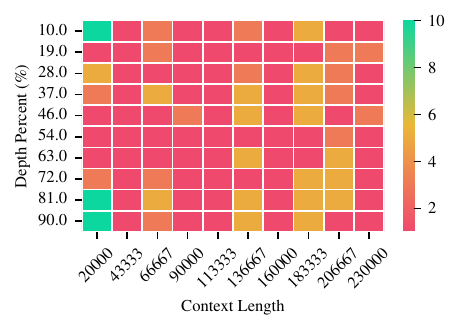}%
        \label{fig:ltrillm_RH_none_Persistent_Voting_none_restrict_needle_within_block}%
    }\hfill
    \subfloat[$\widetilde{\mathcal{RH}},\mathcal{P},\mathcal{V}$]{%
        \includegraphics[width=.24\linewidth]{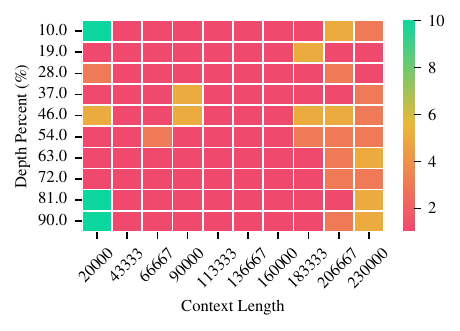}%
        \label{fig:ltrillm_RH_none_Persistent_Voting_restrict_needle_within_block}%
    }\\
    \subfloat[$\mathcal{RH}$, $\widetilde{\mathcal{P}}$, $\widetilde{\mathcal{V}}$]{%
        \includegraphics[width=.24\linewidth]{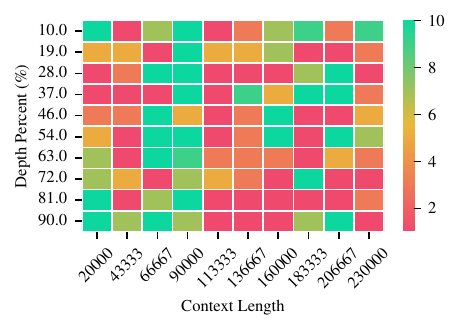}%
        \label{fig:ltrillm_RH_Persistent_none_Voting_none_restrict_needle_within_block}%
    }\hfill
    \subfloat[$\mathcal{RH}$, $\widetilde{\mathcal{P}}$, $\mathcal{V}$]{%
        \includegraphics[width=.24\linewidth]{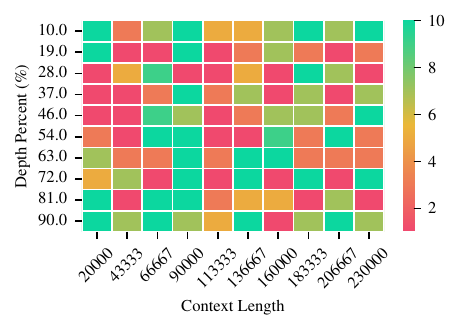}%
        \label{fig:ltrillm_RH_Persistent_none_Voting_restrict_needle_within_block}%
    }\hfill
    \subfloat[$\mathcal{RH}$, $\mathcal{P}$, $\widetilde{\mathcal{V}}$]{%
        \includegraphics[width=.24\linewidth]{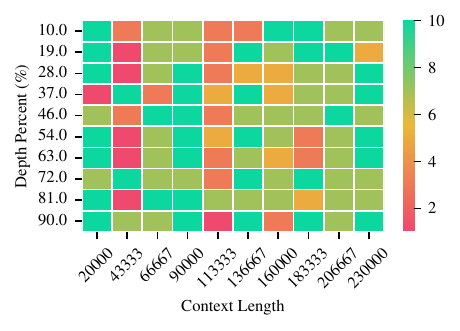}%
        \label{fig:ltrillm_RH_Persistent_Voting_none_restrict_needle_within_block}%
    }\hfill
    \subfloat[$\mathcal{RH},\mathcal{P},\mathcal{V}$]{%
        \includegraphics[width=.24\linewidth]{figures/NIAH-Llama-3-8B-Instruct-262k-LtriLLM-enable_restrict_needle_within_block.pdf}%
        \label{fig:ltrillm_RH_Persistent_Voting_restrict_needle_within_block}%
    }\\
    \caption{Ablation Study of Ltri-LLM on the Persistent, Retrieval Heads and Collaborative Voting Mechanism, represented by $\mathcal{P}$, $\mathcal{RH}$ and $\mathcal{V}$, respectively. The symbol $\widetilde{(\cdot)}$ indicates that the corresponding mechanism is deactivated.}
    \label{fig:ablation}
\end{figure}

\begin{figure}[htbp]
    \subfloat[$\mathcal{Q}=512$]{%
        \includegraphics[width=.33\linewidth]{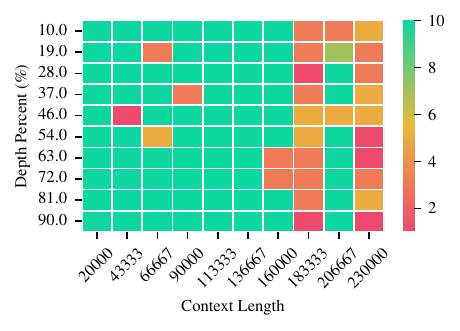}%
        \label{fig:tuning_prefillingLastChunkSize_512}%
    }\hfill
    \subfloat[$\mathcal{Q}=256$]{%
        \includegraphics[width=.33\linewidth]{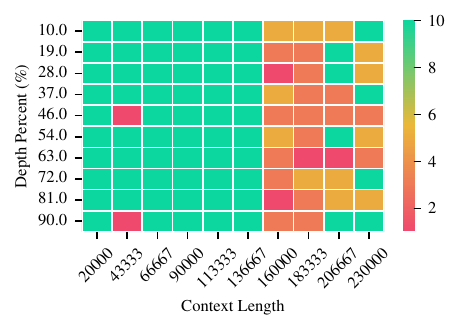}%
        \label{fig:tuning_prefillingLastChunkSize_256}%
    }\hfill
    \subfloat[$\mathcal{Q}=128$]{%
        \includegraphics[width=.33\linewidth]{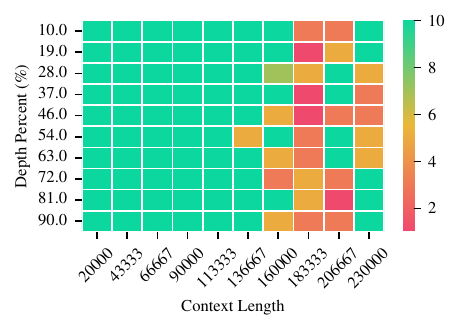}%
        \label{fig:tuning_prefillingLastChunkSize_128}%
    }\\
    \subfloat[$\mathcal{Q}=64$]{%
        \includegraphics[width=.33\linewidth]{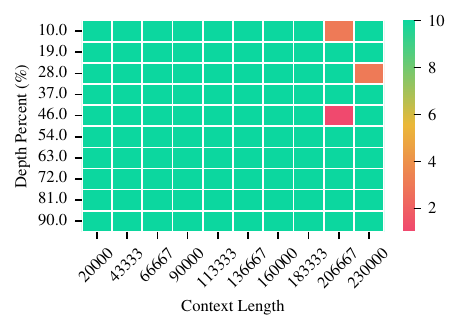}%
        \label{fig:tuning_prefillingLastChunkSize_64}%
    }\hfill
    \subfloat[$\mathcal{Q}=32$]{%
        \includegraphics[width=.33\linewidth]{figures/NIAH-Llama-3-8B-Instruct-262k-LtriLLM-enable_restrict_needle_within_block.pdf}%
        \label{fig:tuning_prefillingLastChunkSize_32}%
    }\hfill
    \subfloat[$\mathcal{Q}=16$]{%
        \includegraphics[width=.33\linewidth]{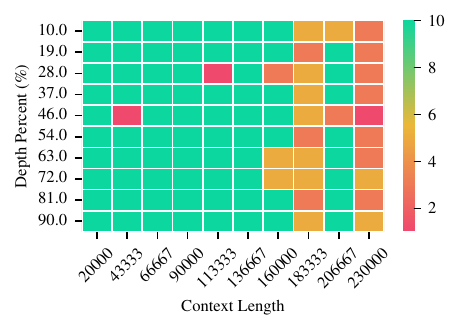}%
        \label{fig:tuning_prefillingLastChunkSize_16}%
    }\\
    \caption{Performance of Ltri-LLM on NIAH for different $\mathcal{Q}$.}
    \label{fig:tuning_prefilling_last_chunk_size}
\end{figure}

\begin{figure}[htbp]
    \subfloat[$\lambda=3$]{%
        \includegraphics[width=.24\linewidth]{figures/NIAH-Llama-3-8B-Instruct-262k-LtriLLM-enable_restrict_needle_within_block.pdf}%
        \label{fig:tuning_lambda_3}%
    }\hfill
    \subfloat[$\lambda=4$]{%
        \includegraphics[width=.24\linewidth]{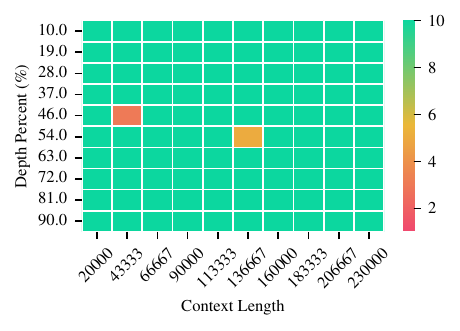}%
        \label{fig:tuning_lambda_4}%
    }\hfill
    \subfloat[$\lambda=5$]{%
        \includegraphics[width=.24\linewidth]{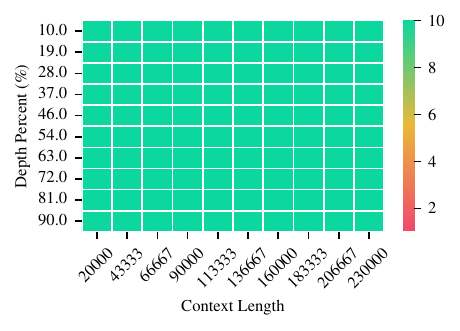}%
        \label{fig:tuning_lambda_5}%
    }\hfill
    \subfloat[$\lambda=6$]{%
        \includegraphics[width=.24\linewidth]{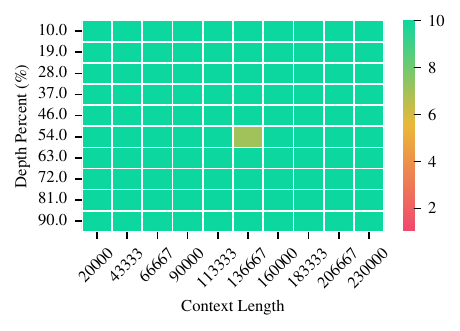}%
        \label{fig:tuning_lambda_6}%
    }\\
    \subfloat[$\lambda=7$]{%
        \includegraphics[width=.24\linewidth]{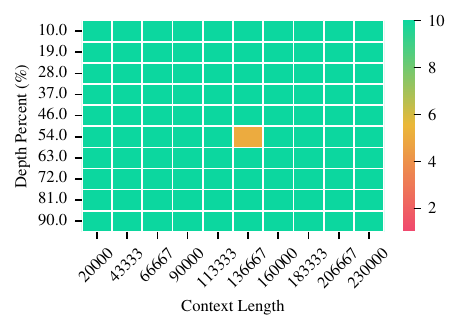}%
        \label{fig:tuning_lambda_7}%
    }\hfill
    \subfloat[$\lambda=8$]{%
        \includegraphics[width=.24\linewidth]{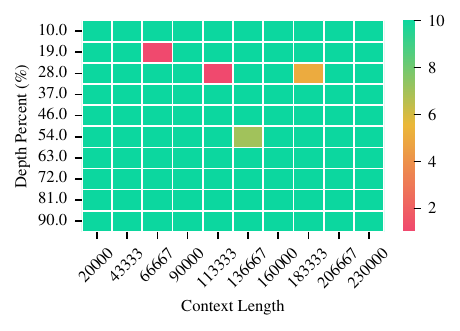}%
        \label{fig:tuning_lambda_8}%
    }\hfill
    \subfloat[$\lambda=9$]{%
        \includegraphics[width=.24\linewidth]{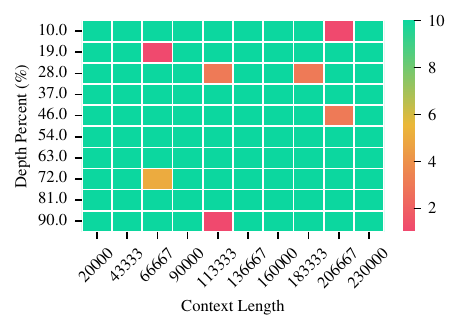}%
        \label{fig:tuning_lambda_9}%
    }\hfill
    \subfloat[$\lambda=10$]{%
        \includegraphics[width=.24\linewidth]{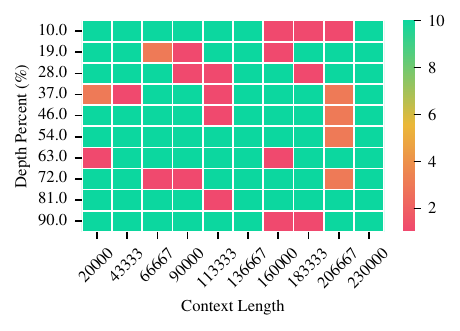}%
        \label{fig:tuning_lambda_10}%
    }
    \caption{Performance of Ltri-LLM on NIAH for different $\lambda$.}
    \label{fig:tuning_lambda_param}
\end{figure}

\begin{figure}[htbp]
    \subfloat[$\Omega=row$]{%
        \includegraphics[width=.33\linewidth]{figures/NIAH-Llama-3-8B-Instruct-262k-LtriLLM-enable_restrict_needle_within_block.pdf}%
        \label{fig:tuning_ratio_row}%
    }\hfill
    \subfloat[$\Omega=col$]{%
        \includegraphics[width=.33\linewidth]{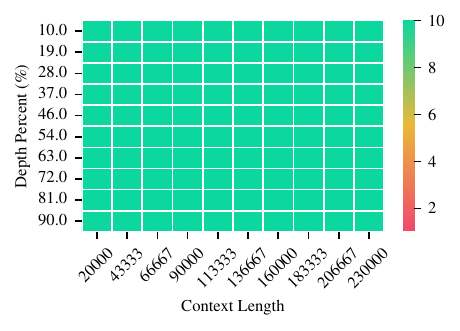}%
        \label{fig:tuning_ratio_col}%
    }\hfill
    \subfloat[$\Omega=rowcol$]{%
        \includegraphics[width=.33\linewidth]{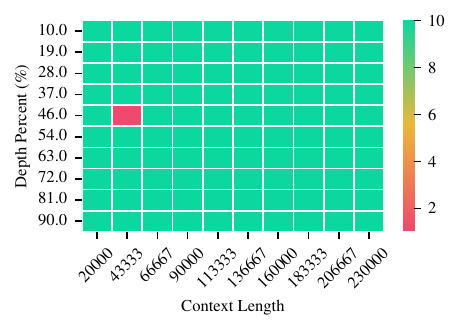}%
        \label{fig:tuning_ratio_rowcol}%
    }
    \caption{Performance of Ltri-LLM on NIAH for different $\Omega$.}
    \label{fig:tuning_ratio_mode}
\end{figure}

\textbf{Hyperparameter Tuning.} 
Initially, we should determine the attention map threshold $\theta$, as described in Section \ref{sec:3.2}, along with the Intersection over Union (IoU) threshold $\varphi$ for the Non-Maximum Suppression (NMS) operator from the torchvision\footnote{https://github.com/pytorch/vision} library. Specifically, $\theta$ and $\phi$ are jointly adjusted via random search for each individual layer to optimize the F1-score of semantic spans overlapping with the evidence. $\theta$ is searched within the interval (0.0001, 0.95) uniformly, while $\varphi$ is explored uniformly within the interval (0.01, 0.95). Detailed values of $\theta$ and $\varphi$ can be found in Appendix \ref{appendix::span_division_nms}. 
The typical reference values for $\theta$ and $\varphi$ are 90$^{th}$ percentile and 0.1, respectively. 
For the prefilling last chunk size $\mathcal{Q}$, we vary $\mathcal{Q}$ in $\{16, 32, 64, 128, 256, 512\}$ and find the best performance is achieved when $\mathcal{Q}=32$ as shown in Figure \ref{fig:tuning_prefilling_last_chunk_size}. 
Next, we fix the optimal $\mathcal{Q}$ and report the best performance for various $\lambda$ values in $\{3, 4, 5, 6, 7, 8, 9, 10\}$. Figure \ref{fig:tuning_lambda_param} suggests that Ltri-LLM keeps decent performance for smaller values of $\lambda$, while the performance deteriorates when $\lambda$ becomes larger. 
Finally, we fix the best $\mathcal{Q}$ and $\lambda$ values and explore different ratio modes $\Omega$ in Figure \ref{fig:tuning_ratio_mode}. The results indicate that all ratio calculation methods, i.e., row, col, and rowcol can yield ideal performance.

\begin{figure}[htbp]
\begin{center}
\centerline{\includegraphics[width=\columnwidth]{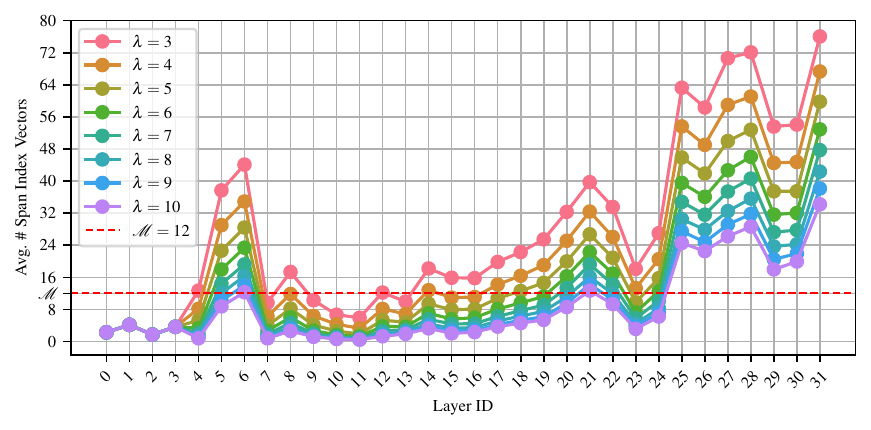}}
\caption{Average Number of Span Index Vectors for different $\lambda$.}
\label{fig:avg_span_index_vectors}
\end{center}
\end{figure}

\textbf{Compression Ratio.} For each block of length $\mathcal{B}=128$, up to $\mathcal{M}=12$ span index vectors can be generated. Figure \ref{fig:avg_span_index_vectors} shows the average number of span index vectors generated per block at all layers for various $\lambda$ values. We set $\lambda=20$ uniformly for the first four layers. Generally, the average number of span index vectors per block increases with layer depth, reflecting a larger ratio value and aligning with the triangular attention pattern observed in Figure \ref{fig:triangle_attention_pattern}. Additionally, as $\lambda$ rises, the average number of span index vectors per block decreases, consistent with Equation \ref{eq:rv}. For a sequence length $\mathcal{L}$, let $\mathcal{H}$ be the number of attention heads and $h$ the number of applied retrieval heads; Ltri-LLM will approximately retain $\mathcal{L}/\mathcal{B}*\mathcal{M}*h$ span index vectors after prefilling. A relaxed lower bound for the compression ratio $\delta$ is $\frac{\mathcal{B}*\mathcal{H}}{\mathcal{M}*h}$.

\begin{table*}[htbp]
\centering
\caption{Occupied GPU memory space (GiB) of evicted tokens for different methods evaluated at lengths from 100K to 800K.}
\resizebox{0.7\textwidth}{!}{
\begin{NiceTabular}{lcccccccc}
\CodeBefore
  \rowcolors{2}{white}{lightgray!20}
\Body
\toprule
        Methods & 100K & 200K & 300K & 400K & 500K & 600K & 700K & 800K \\
        \midrule
        LLAMA3-8B-Instruct-262K & 23.344 & 47.781 & 72.188 & 96.594 & 121.000 & 145.438 & 169.844 & 194.250 \\
        InfLLM-reprTopk12 & 2.241 & 4.587 & 6.930 & 9.273 & 11.616 & 13.962 & 16.305 & 18.648 \\
        Ltri-LLM$_{\mathcal{M}=12,\lambda=3}$ & 0.027 & 0.057 & 0.087 & 0.117 & 0.148 & 0.178 & 0.209 & 0.240 \\
        Ltri-LLM$_{\mathcal{M}=12,\lambda=4}$ & 0.025 & 0.052 & 0.081 & 0.110 & 0.139 & 0.168 & 0.198 & 0.227 \\
        Ltri-LLM$_{\mathcal{M}=12,\lambda=5}$ & 0.022 & 0.046 & 0.072 & 0.098 & 0.125 & 0.152 & 0.180 & 0.207 \\
        Ltri-LLM$_{\mathcal{M}=12,\lambda=6}$ & 0.019 & 0.040 & 0.063 & 0.086 & 0.111 & 0.134 & 0.159 & 0.184 \\
        Ltri-LLM$_{\mathcal{M}=12,\lambda=7}$ & 0.017 & 0.035 & 0.055 & 0.077 & 0.098 & 0.119 & 0.141 & 0.163 \\
        Ltri-LLM$_{\mathcal{M}=12,\lambda=8}$ & 0.015 & 0.031 & 0.048 & 0.067 & 0.086 & 0.105 & 0.125 & 0.144 \\
        Ltri-LLM$_{\mathcal{M}=12,\lambda=9}$ & 0.013 & 0.027 & 0.043 & 0.060 & 0.077 & 0.094 & 0.111 & 0.129 \\
        Ltri-LLM$_{\mathcal{M}=12,\lambda=10}$ & 0.012 & 0.024 & 0.039 & 0.054 & 0.069 & 0.084 & 0.100 & 0.116 \\
        \bottomrule
\end{NiceTabular}
}
\label{tab:block_k_bytes}
\end{table*}

\section{Related Work}
\textbf{Scaling Up Context Window of LLMs.} The primary strategy for scaling up the context window of LLMs involves adjusting the parameters of position embeddings like Rotary Position Embedding (RoPE) \citep{su2021roformer}, including training-free methods \citep{chen2023extending,dynamicNTK,peng2023yarn,liu2024scaling}, and continual-training methods \citep{rozière2024codellamaopenfoundation,fu2024data,bai2024longalignrecipelongcontext}. While these methods show promising results in long context tasks, the quadratic complexity of Full-Attention (FA) imposes heavy burdens on memory and inference time which greatly restricts the real-world application of long context LLMs.

\textbf{Attention Pattern Analysis}
Streaming-LLM \citep{xiao2023streamingllm} and LM-Infinite \citep{han2023lm}  identified the attention sink phenomenon, noting that preserving initial tokens aids in restoring the performance of sliding window attention. 
FlexGen \citep{ge2024modeltellsdiscardadaptive} discovered attention heads usually have different structures and propose four KV cache compression policies. MInference \citep{jiang2024minference} summarized three unique patterns in long context attention matrices and directly adopt sparse attention kernels in the pre-filling stage. PyramidKV \citep{cai2024pyramidkvdynamickvcache} discovered that shallow layers show more localized attention and deep layers show more massive activation and it  assigns more KV cache to shallow layers and less to deep ones.

\textbf{Long Context Inference Acceleration}
One kind of the long context inference acceleration method is to reduce the memory space of KV while keep the computation complexity of attention unchanged, including reducing the hidden states space  \citep{deepseekai2024deepseekv2strongeconomicalefficient,ainslie2023gqatraininggeneralizedmultiquery,hooper2024kvquant} and share KV across layers  \citep{sun2024cacheoncedecoderdecoderarchitectures,brandon2024reducingtransformerkeyvaluecache}. Another kind is to reduce the space and computation simultaneously on the sequence dimension. Some of them involve additional training  \citep{nawrot2024dynamicmemorycompressionretrofitting,munkhdalai2024leavecontextbehindefficient,yang2024textmemory3languagemodelingexplicit} or another structure like State Space Models (SSMs) \citep{gu2024mambalineartimesequencemodeling,lieber2024jambahybridtransformermambalanguage,dao2024transformersssmsgeneralizedmodels}. In this paper, we mainly focus on training-free acceleration methods, including dynamic sparse attention  \citep{jiang2024minference,tang2024questqueryawaresparsityefficient,ribar2024sparqattentionbandwidthefficientllm,yang2024posttrainingsparseattentiondouble} and KV compression methods \citep{zhang2023h2oheavyhitteroracleefficient,li2024snapkv,lee2024infinigenefficientgenerativeinference,liu2024retrievalattentionacceleratinglongcontextllm,ge2024modeltellsdiscardadaptive,adnan2024keyformerkvcachereduction,cai2024pyramidkvdynamickvcache,xiao2024infllm}. 

\section{Conclusion}

In this paper, we discover the inference obstacles of InfLLM by presenting the strong correlation between reasonable model responses and accurate evidence recall. Based on the observation of triangular LLM attention distribution, we propose Ltri-LLM, a novel approach that efficiently and flexibly identifies semantic spans within long contexts. Extensive experiments on NIAH, $\infty$-Bench, and RULER verify the effectiveness of Ltri-LLM.

\bibliography{icml2025}
\bibliographystyle{icml2025}

\newpage
\appendix
\onecolumn
\section{Details of Mandatory Evidence Injection}\label{appendix::fan}
\begin{figure}[htbp]
    \centering
    \includegraphics[width=\textwidth]{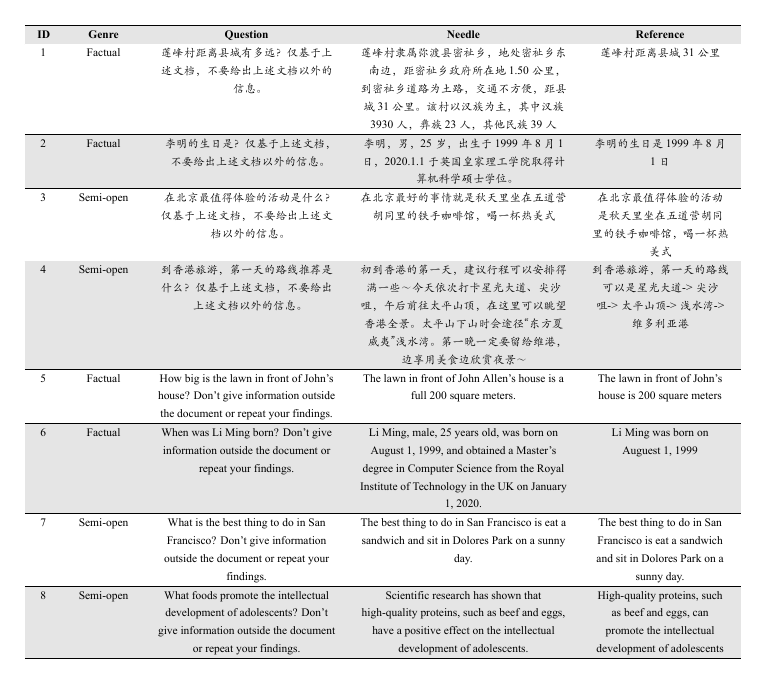}
    \caption{Needle-In-A-Haystack benchmarks}
    \label{fig:niah}
\end{figure}

Besides the original NIAH test, we construct other seven additional bilingual NIAH tasks, as shown in Figure \ref{fig:niah}. including factual and semi-open genres.
For each task, we evaluate LLM with 35 context lengths evenly distributed from 10K to 230K. The insertion position of needle is uniformly varied from 0\% to 100\% for each context length, resulting in a total combination of 1,225 tests.

\begin{figure}[htbp]
    \subfloat[LLAMA3-8B-Instruct-262K, InfLLM, Needle-1]{%
        \includegraphics[width=.48\linewidth]{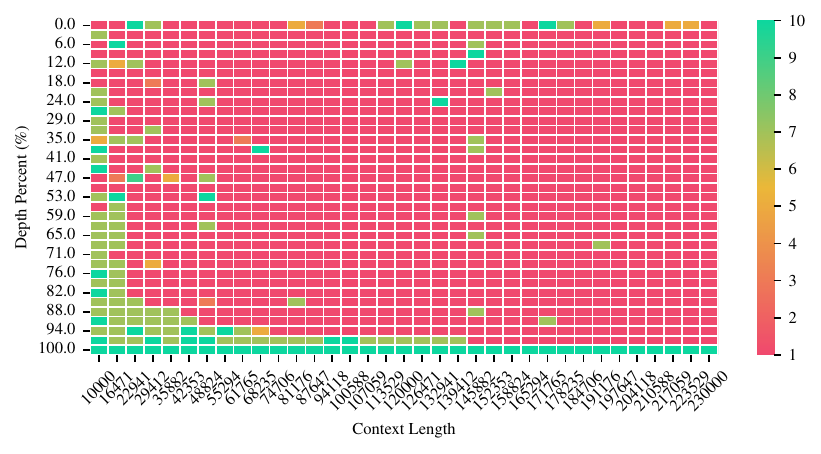}%
        \label{fig:l38i262-infllm-custom_niah-needle_1}%
    }\hfill
    \subfloat[LLAMA3-8B-Instruct-262K, InfLLM, Needle-2]{%
        \includegraphics[width=.48\linewidth]{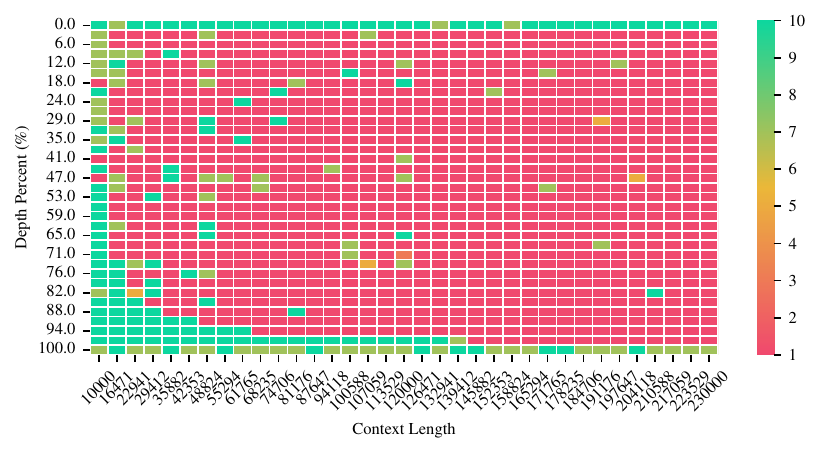}%
        \label{fig:l38i262-infllm-custom_niah-needle_2}%
    }\\
    \subfloat[LLAMA3-8B-Instruct-262K, InfLLM, Needle-3]{%
        \includegraphics[width=.48\linewidth]{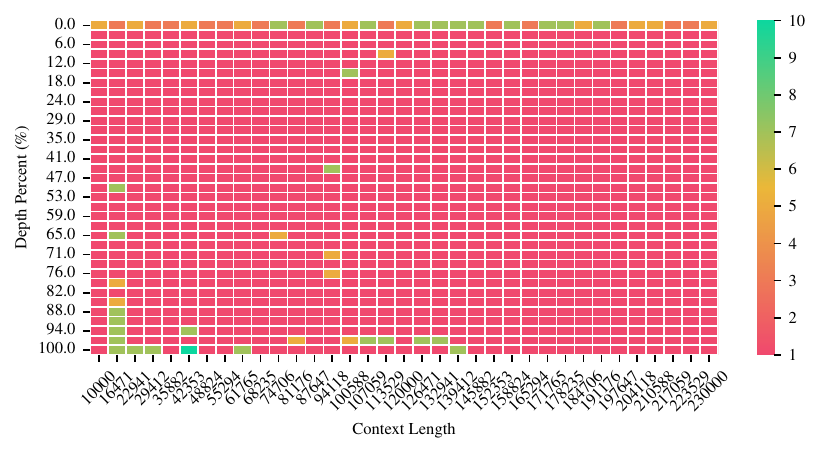}%
        \label{fig:l38i262-infllm-custom_niah-needle_3}%
    }\hfill
    \subfloat[LLAMA3-8B-Instruct-262K, InfLLM, Needle-4]{%
        \includegraphics[width=.48\linewidth]{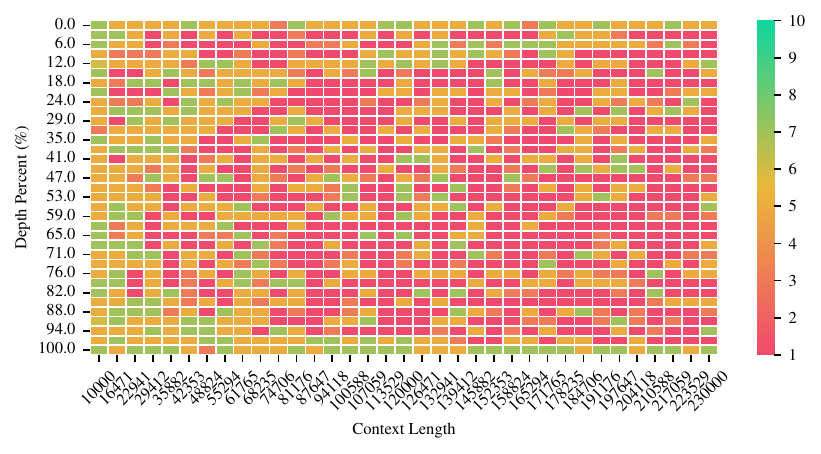}%
        \label{fig:l38i262-infllm-custom_niah-needle_4}%
    }\\
    \subfloat[LLAMA3-8B-Instruct-262K, InfLLM, Needle-5]{%
        \includegraphics[width=.48\linewidth]{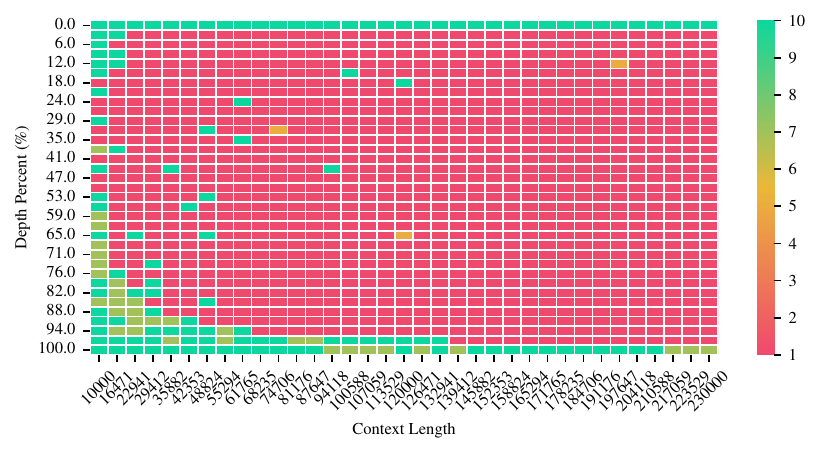}%
        \label{fig:l38i262-infllm-custom_niah-needle_5}%
    }\hfill
    \subfloat[LLAMA3-8B-Instruct-262K, InfLLM, Needle-6]{%
        \includegraphics[width=.48\linewidth]{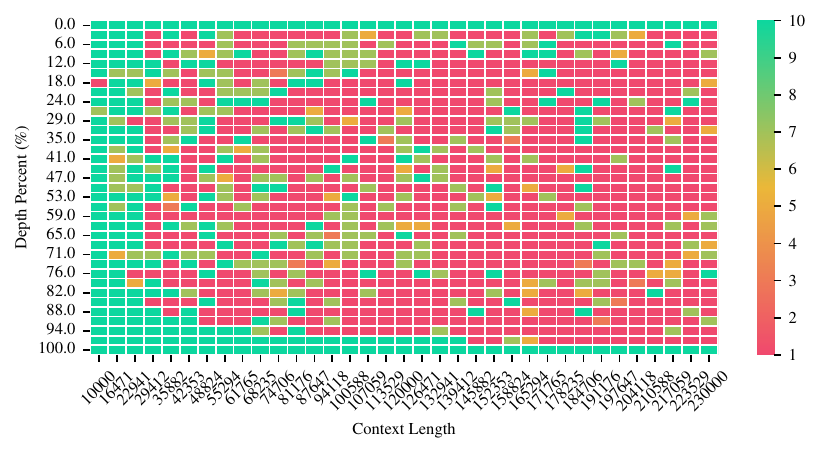}%
        \label{fig:l38i262-infllm-custom_niah-needle_6}%
    }\\
    \subfloat[LLAMA3-8B-Instruct-262K, InfLLM, Needle-7]{%
        \includegraphics[width=.48\linewidth]{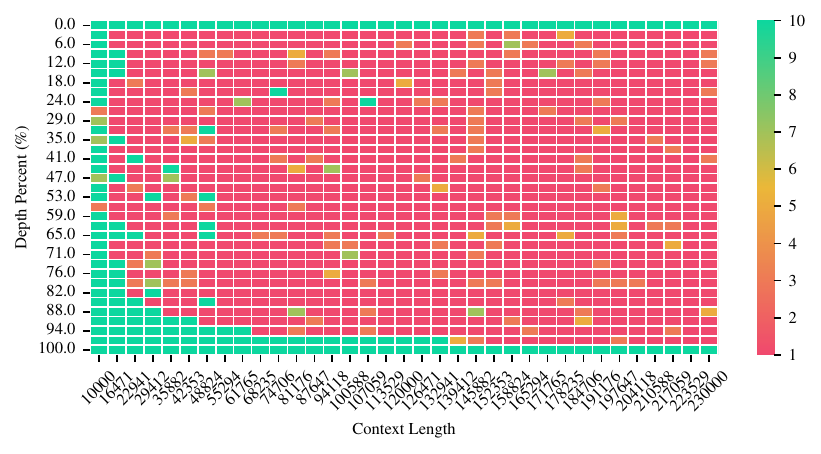}%
        \label{fig:l38i262-infllm-custom_niah-needle_7}%
    }\hfill
    \subfloat[LLAMA3-8B-Instruct-262K, InfLLM, Needle-8]{%
        \includegraphics[width=.48\linewidth]{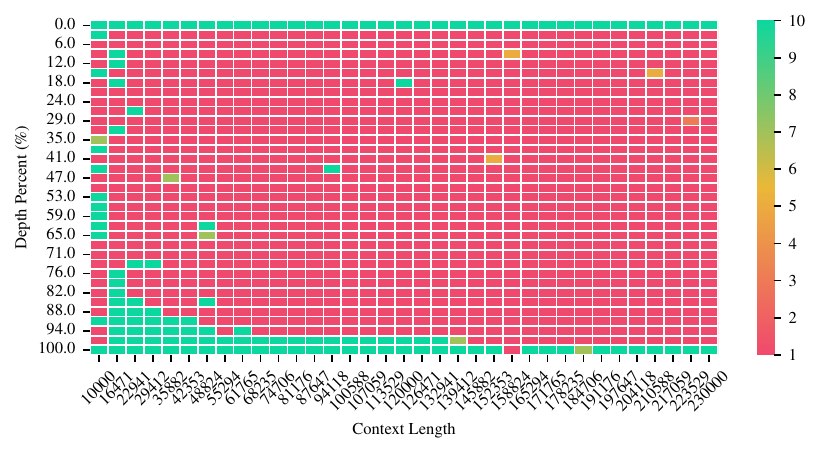}%
        \label{fig:l38i262-infllm-custom_niah-needle_8}%
    }\\
    \caption{LLAMA3-8B-Instruct-262K-InfLLM on NIAH benchmarks.}
    \label{fig:l38i262-infllm-custom_niah}
\end{figure}

\begin{figure}[htbp]
    \subfloat[InternLM-2-Chat-7B, InfLLM, Needle-1]{%
        \includegraphics[width=.48\linewidth]{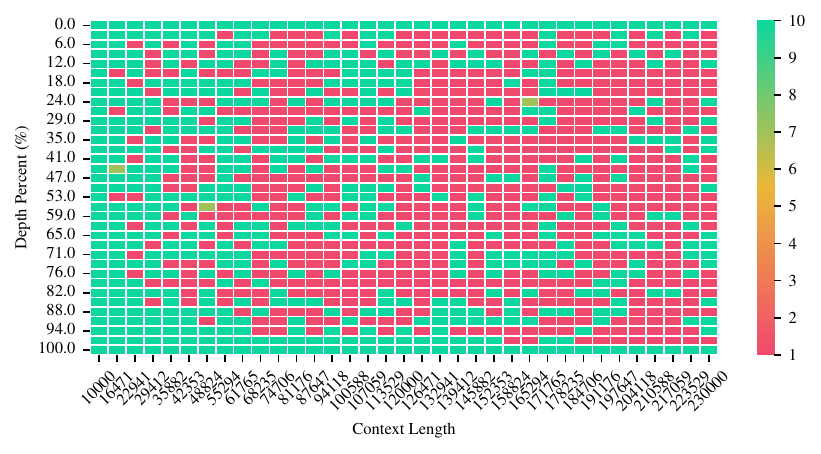}%
        \label{fig:i2c7-infllm-custom_niah-needle_1}%
    }\hfill
    \subfloat[InternLM-2-Chat-7B, InfLLM, Needle-2]{%
        \includegraphics[width=.48\linewidth]{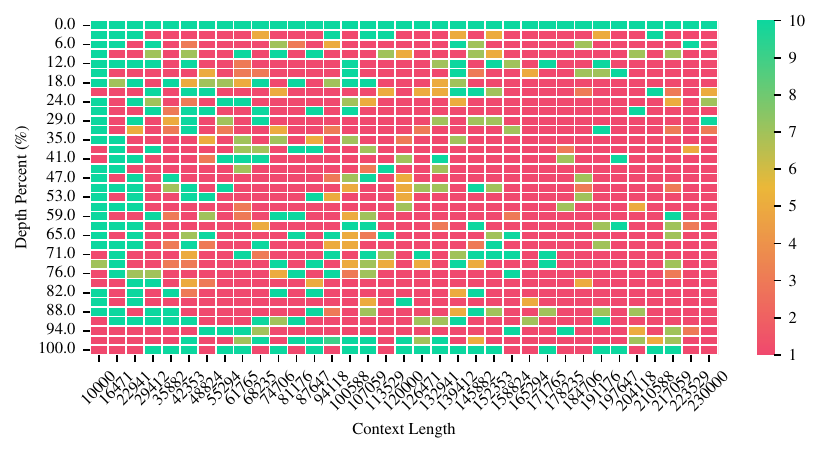}%
        \label{fig:i2c7-infllm-custom_niah-needle_2}%
    }\\
    \subfloat[InternLM-2-Chat-7B, InfLLM, Needle-3]{%
        \includegraphics[width=.48\linewidth]{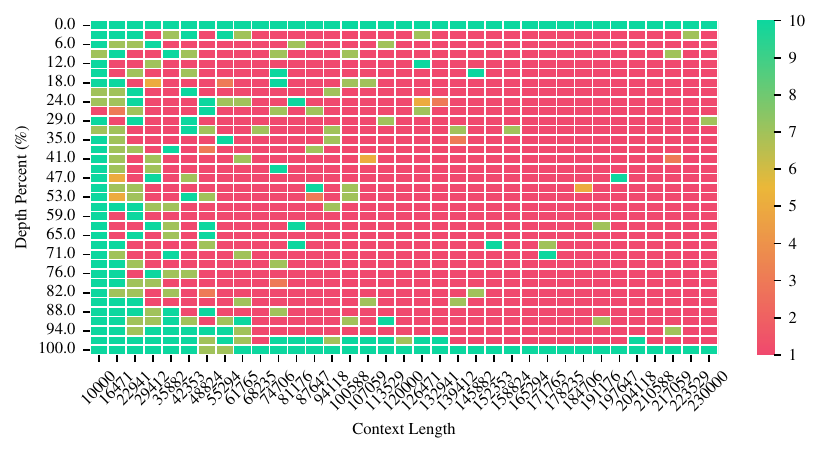}%
        \label{fig:i2c7-infllm-custom_niah-needle_3}%
    }\hfill
    \subfloat[InternLM-2-Chat-7B, InfLLM, Needle-4]{%
        \includegraphics[width=.48\linewidth]{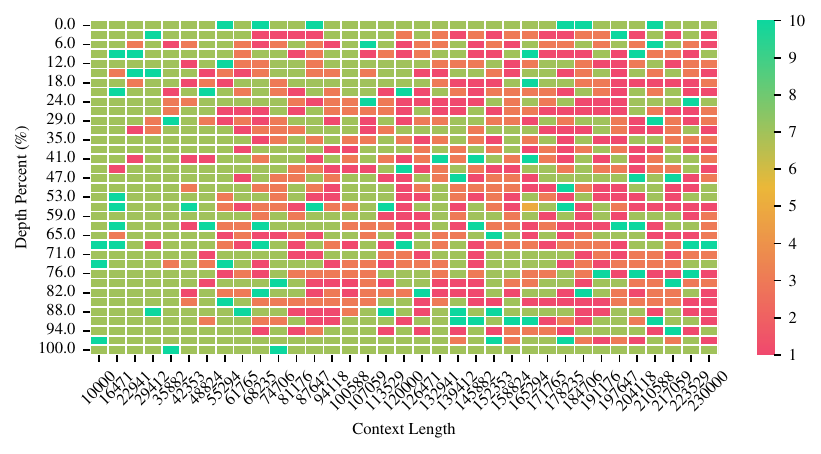}%
        \label{fig:i2c7-infllm-custom_niah-needle_4}%
    }\\
    \subfloat[InternLM-2-Chat-7B, InfLLM, Needle-5]{%
        \includegraphics[width=.48\linewidth]{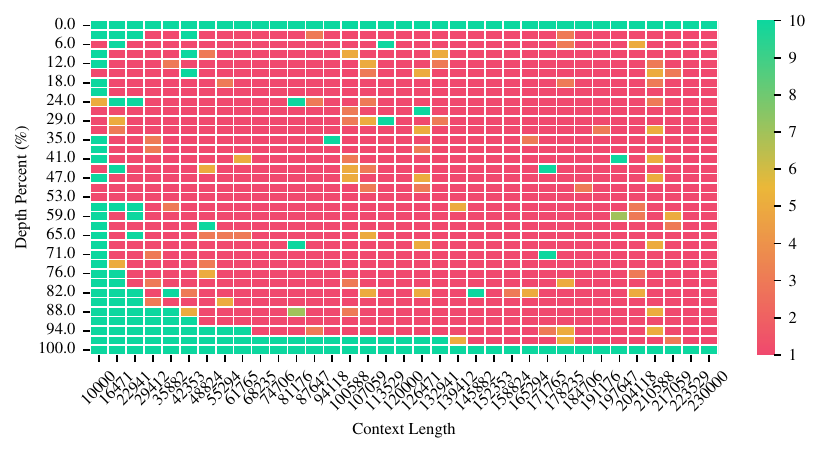}%
        \label{fig:i2c7-infllm-custom_niah-needle_5}%
    }\hfill
    \subfloat[InternLM-2-Chat-7B, InfLLM, Needle-6]{%
        \includegraphics[width=.48\linewidth]{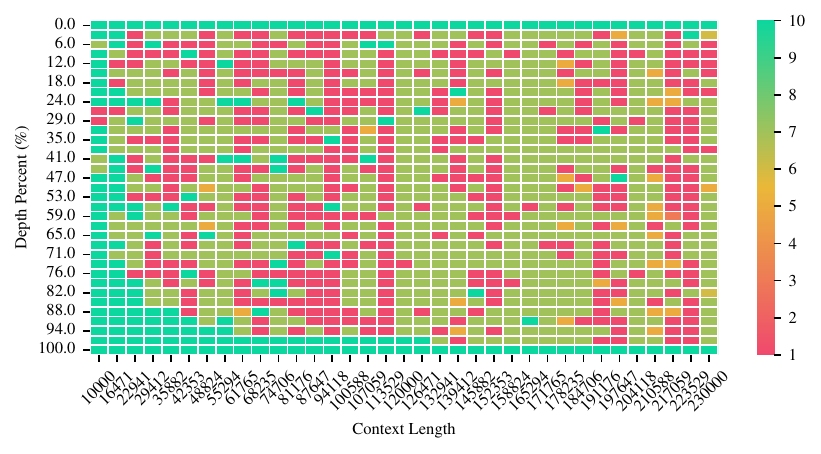}%
        \label{fig:i2c7-infllm-custom_niah-needle_6}%
    }\\
    \subfloat[InternLM-2-Chat-7B, InfLLM, Needle-7]{%
        \includegraphics[width=.48\linewidth]{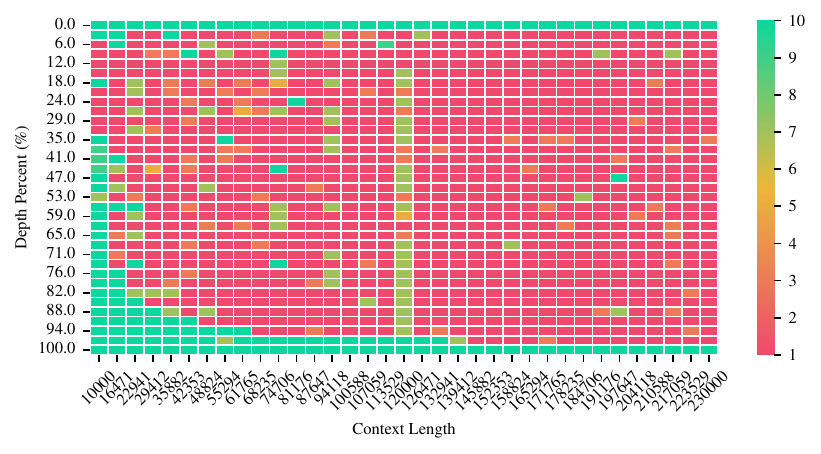}%
        \label{fig:i2c7-infllm-custom_niah-needle_7}%
    }\hfill
    \subfloat[InternLM-2-Chat-7B, InfLLM, Needle-8]{%
        \includegraphics[width=.48\linewidth]{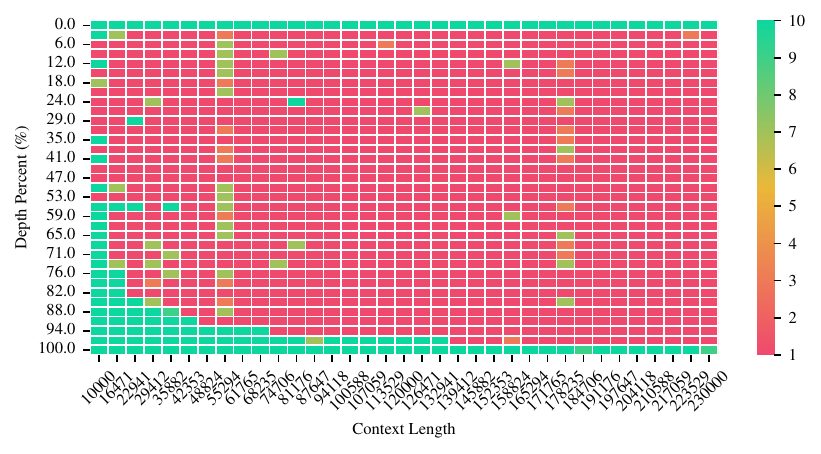}%
        \label{fig:i2c7-infllm-custom_niah-needle_8}%
    }\\
    \caption{InternLM-2-Chat-7B-InfLLM on NIAH benchmarks.}
    \label{fig:i2c7-infllm-custom_niah}
\end{figure}

Figure \ref{fig:l38i262-infllm-custom_niah} and \ref{fig:i2c7-infllm-custom_niah} illustrate the performance of LLAMA3-8B-Instruct-262K and InternLM-2-Chat-7B with InfLLM across all NIAH tasks as depicted in Figure \ref{fig:niah}. Generally, InfLLM's design allows the LLM direct evidence access during the decoding stage when the evidence is inserted at the very beginning or end of the prompt, leading to superior performance. Subsequently, we choose several unsuccessful combinations from each NIAH task, and activate the mandatory evidence injection mechanism during their second execution to observe whether the model's response shifts from failure to success.

\begin{figure}[htbp]
    \subfloat[LLAMA3-8B-Instruct-262K]{%
        \includegraphics[width=.48\linewidth]{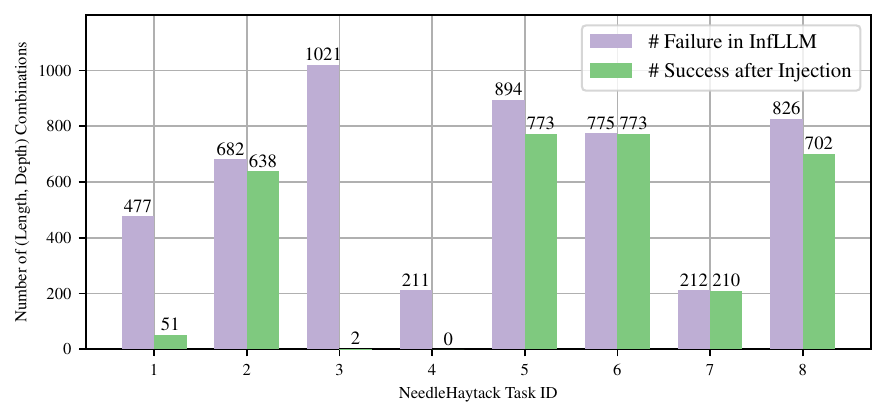}%
        \label{fig:fan_llama3_8b_instruct_262k}%
    }\hfill
    \subfloat[InternLM2-Chat-7B]{%
        \includegraphics[width=.48\linewidth]{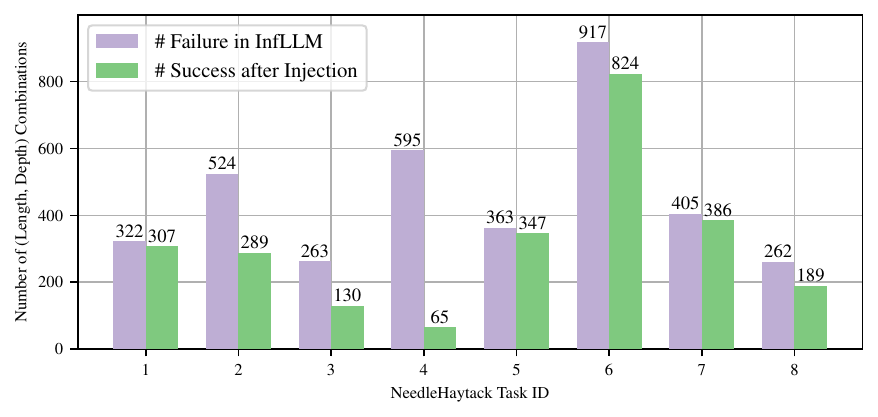}%
        \label{fig:fan_internlm2_chat_7b}%
    }
    \caption{Performance of InfLLM on NIAH benchmarks w/ Mandatory Evidence Injection.}
    \label{fig:fan_l3bi262_i2c7}
\end{figure}

The experimental results depicted in Figure \ref{fig:fan_llama3_8b_instruct_262k} indicate that the approach of mandatory evidence injection into $\mathcal{C}$ effectively convert most failure cases from LLAMA3-8B-Instruct-262K equipped with InfLLM into successful ones across all English NIAH benchmarks, while the similar results merely occur on the second Chinese NIAH test. The phenomenon is attributed to two primary factors: (1) The training corpus for LLAMA3-8B-Instruct-262K is predominantly English, contributing to exceptional performance on English benchmarks.
While InternLM2-Chat-7B \citep{cai2024internlm2}, undergoing extensive pre-training on Chinese corpora, demonstrates observable conversion across both Chinese and English needle-in-a-haystack tests, as shown in Figure \ref{fig:fan_internlm2_chat_7b}.
(2) The model response is deemed valid only if it achieves the highest score in the GPT-4 evaluation. Nevertheless, significant improvements in the model responses have been observed following the mandatory evidence injection sometimes, even if they have not met the success criterion yet.

\begin{wrapfigure}{r}{0.5\columnwidth}
    \vspace{-15pt}
    \centering
    \includegraphics[width=0.65\linewidth]{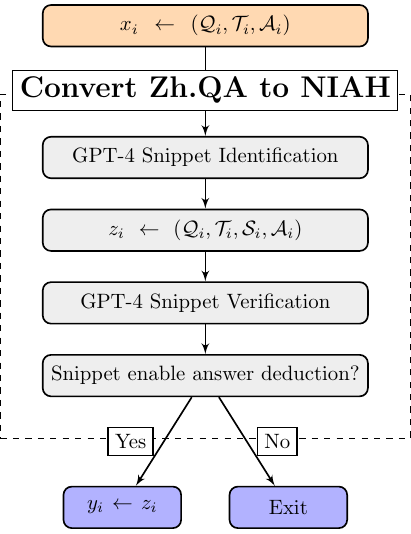}
    \caption{Convert Zh.QA dataset from $\infty$-Bench benchmark into NIAH format. $\mathcal{Q}, \mathcal{T}, \mathcal{A}$, and $\mathcal{S}$ stand for retrieval question, passage, reference answer, and snippet with needle, respectively.}
    \label{fig:zhqa_into_niah}
\end{wrapfigure}
\sethlcolor{gitred}
To further validate the proposition in a more realistic scenario, we also examine the conclusion in a Chinese Question Answering dataset (Zh.QA) from InfiniteBench, which includes 175 examples with the average token length of more than 2,000 thousand. However, Zh.QA does not provide evidences corresponding to the answers. To locate the evidences, we conduct a two-round dialogue with GPT-4, as shown in Figure \ref{fig:zhqa_into_niah}. In the first round, we require GPT-4 to identify the exact snippet in the passage that corresponds to the reference answer for the given question. In the second round, we let GPT-4 determine whether the information in the snippet allows for an accurate deduction of the answer to the question. After these two rounds of dialogue, we end up with a reliable converted Zh.QA dataset with the NIAH format.

There are a total of 65 pieces of data with context length less than 1M tokens in the Zh.QA dataset. When using the LLAMA3-8B-Instruct-262K model with InfLLM framework for inference, there are 12 pieces of data whose model response get the highest score after GPT-4 evaluation. Through the mandatory evidence injection operation, the model response for 28 pieces of data are considered valid in the GPT-4 evaluation.

\newpage
\section{Experimental Details}\label{appendix::configuration}
Table \ref{tab:configuration} presents the specific hyperparameter configurations for various tested benchmarks. As highlighted in Section \ref{sec::ablation_and_analysis}, the prefilling last chunk size $\mathcal{Q}$ significantly impacts the performance, thus we calibrate $\mathcal{Q}$ to appropriate values for different benchmarks. The attention map threshold $\theta$ and Intersection over Union (IoU) threshold $\varphi$ for the Non-Maximum Suppression (NMS) algorithm strike a balance semantic span discovery and noise inclusion, as discussed in Appendix \ref{appendix::span_division_nms}. To establish a more universal set of $\theta$ and $\varphi$ values on a larger synthetic dataset is left for future work. It's recommended to refer to Section \ref{sec::ablation_and_analysis} for the meanings and effects of notations $\Omega$, $\lambda$ and $\mathcal{M}$.

\begin{table}[htbp]
\centering
\caption{Hyperparameter configurations of Ltri-LLM for different benchmarks.}
\vskip 0.15in
\begin{small}
\begin{NiceTabular}{lcccccccc}
\CodeBefore
  \rowcolors{2}{white}{lightgray!20}
\Body
\toprule
Benchmark & Mask & $\Omega$ & $\lambda$ & Span & $\mathcal{M}$ & $\theta$ & $\varphi$ & $\mathcal{Q}$\\
\midrule
\texttt{NIAH} & $\triangle$ & row & 3 & 4 & 12 & random search & random search & 32 \\
\midrule
$\infty$-Bench/En.Sum & $\triangle$ & row & 3 & 4 & 12 & random search & random search & 32 \\
$\infty$-Bench/Zh.QA & $\triangle$ & row & 3 & 4 & 12 & random search & random search & 64 \\
$\infty$-Bench/En.QA & $\triangle$ & row & 3 & 4 & 12 & random search & random search & 48 \\
$\infty$-Bench/En.MC & $\triangle$ & row & 3 & 4 & 12 & random search & random search & 256 \\
$\infty$-Bench/En.Dia & $\triangle$ & row & 3 & 4 & 12 & random search & random search & 48 \\
$\infty$-Bench/Code.Debug & $\triangle$ & row & 3 & 4 & 12 & random search & random search & 256 \\
$\infty$-Bench/Math.Find & $\triangle$ & row & 3 & 4 & 12 & random search & random search & 32 \\
$\infty$-Bench/Retr.PassKey & $\triangle$ & row & 3 & 4 & 12 & random search & random search & 32 \\
$\infty$-Bench/Retr.Number & $\triangle$ & row & 3 & 4 & 12 & random search & random search & 32 \\
$\infty$-Bench/Retr.KV & $\triangle$ & row & 3 & 4 & 12 & random search & random search & 48 \\
\midrule
RULER/niah\_single\_1 & $\triangle$ & row & 3 & 6 & 24 & $90^{\text{th}} \text{ percentile}$ & 0.01 & 48 \\
RULER/niah\_single\_2 & $\triangle$ & row & 3 & 6 & 24 & $90^{\text{th}} \text{ percentile}$ & 0.01 & 48 \\
RULER/niah\_single\_3 & $\triangle$ & row & 3 & 6 & 24 & $90^{\text{th}} \text{ percentile}$ & 0.01 & 40 \\
RULER/niah\_multikey\_1 & $\triangle$ & row & 3 & 6 & 24 & $90^{\text{th}} \text{ percentile}$ & 0.01 & 40 \\
RULER/niah\_multikey\_2 & $\triangle$ & row & 3 & 6 & 24 & $90^{\text{th}} \text{ percentile}$ & 0.01 & 40 \\
RULER/niah\_multikey\_3 & $\triangle$ & row & 3 & 6 & 24 & $90^{\text{th}} \text{ percentile}$ & 0.01 & 96 \\
RULER/niah\_multivalue & $\triangle$ & row & 3 & 6 & 24 & $90^{\text{th}} \text{ percentile}$ & 0.01 & 40 \\
RULER/niah\_multiquery & $\triangle$ & row & 3 & 6 & 24 & $90^{\text{th}} \text{ percentile}$ & 0.01 & 80 \\
RULER/vt & $\triangle$ & row & 3 & 6 & 24 & $90^{\text{th}} \text{ percentile}$ & 0.01 & 64 \\
RULER/cwe & $\triangle$ & row & 3 & 6 & 24 & $90^{\text{th}} \text{ percentile}$ & 0.01 & 36 \\
RULER/fwe & $\triangle$ & row & 3 & 6 & 24 & $90^{\text{th}} \text{ percentile}$ & 0.01 & 48 \\
RULER/qa\_1 & $\triangle$ & row & 3 & 6 & 24 & $90^{\text{th}} \text{ percentile}$ & 0.01 & 24 \\
RULER/qa\_2 & $\triangle$ & row & 3 & 6 & 24 & $90^{\text{th}} \text{ percentile}$ & 0.01 & 36 \\
\bottomrule
\end{NiceTabular}
\label{tab:configuration}
\end{small}
\vskip -0.1in
\end{table}

The involved random search results for attention map threshold $\theta$ and Intersection over Union (IoU) threshold $\varphi$ for the Non-Maximum Suppression (NMS) operator from the torchvision as follows:

\begin{table}[H]
    \fontsize{5.8}{5}\selectfont
    \centering
	\begin{minipage}{0.44\textwidth}
        \caption{Attention Map Threshold $\theta$ ($\times 100\%  \text{quantile}$).}
        \label{tab:theta_values}
        \centering
        \resizebox{\textwidth}{!}{
        \begin{NiceTabular}{cccc}
        \CodeBefore
          \rowcolors{2}{white}{lightgray!20}
        \Body
            \toprule
            \textit{\textbf{Layer-0}} & \textit{\textbf{Layer-1}} & \textit{\textbf{Layer-2}} & \textit{\textbf{Layer-3}} \\
            \midrule
            0.9311282274784112 & 0.9080013962575452 & 0.9261057430192952 & 0.9039131198271712 \\
            \midrule
            \textit{\textbf{Layer-4}} & \textit{\textbf{Layer-5}} & \textit{\textbf{Layer-6}} & \textit{\textbf{Layer-7}} \\
            \midrule
            0.930722988460956 & 0.8813112880282284 & 0.89366157731785 & 0.9287055121351976 \\
            \midrule
            \textit{\textbf{Layer-8}} & \textit{\textbf{Layer-9}} & \textit{\textbf{Layer-10}} & \textit{\textbf{Layer-11}} \\
            \midrule
            0.9063453252976758 & 0.9177581400806004 & 0.9306567072342092 & 0.9276351205222796 \\
            \midrule
            \textit{\textbf{Layer-12}} & \textit{\textbf{Layer-13}} & \textit{\textbf{Layer-14}} & \textit{\textbf{Layer-15}} \\
            \midrule
            0.9242215766864216 & 0.9213086907256328 & 0.916481440773853 & 0.924288489083868 \\
            \midrule
            \textit{\textbf{Layer-16}} & \textit{\textbf{Layer-17}} & \textit{\textbf{Layer-18}} & \textit{\textbf{Layer-19}} \\
            \midrule
            0.9254682715409672 & 0.927788224471432 & 0.9131444756355042 & 0.9187994563658362 \\
            \midrule
            \textit{\textbf{Layer-20}} & \textit{\textbf{Layer-21}} & \textit{\textbf{Layer-22}} & \textit{\textbf{Layer-23}} \\
            \midrule
            0.9098651028870453 & 0.9193856460842624 & 0.9049153144323192 & 0.9362935214356204 \\
            \midrule
            \textit{\textbf{Layer-24}} & \textit{\textbf{Layer-25}} & \textit{\textbf{Layer-26}} & \textit{\textbf{Layer-27}} \\
            \midrule
            0.916154415186644 & 0.9062840706971326 & 0.9090799581259756 & 0.8846966017709177 \\
            \midrule
            \textit{\textbf{Layer-28}} & \textit{\textbf{Layer-29}} & \textit{\textbf{Layer-30}} & \textit{\textbf{Layer-31}} \\
            \midrule
            0.9101323777522524 & 0.924574279333147 & 0.9095718506155552 & 0.9473738506727576 \\
            \bottomrule
        \end{NiceTabular}
        }
	\end{minipage}
	\hfill
	\begin{minipage}{0.47\textwidth}
		\centering
        \caption{Intersection over Union Threshold $\varphi$.}
        \label{tab:iou_values}
        \resizebox{\textwidth}{!}{
        \begin{NiceTabular}{cccc}
        \CodeBefore
          \rowcolors{2}{white}{lightgray!20}
        \Body
            \toprule
            \textit{\textbf{Layer-0}} & \textit{\textbf{Layer-1}} & \textit{\textbf{Layer-2}} & \textit{\textbf{Layer-3}} \\
            \midrule
            0.032575607787946555 & 0.11662581959521869 & 0.024889873774503527 & 0.03588605266631541 \\
            \midrule
            \textit{\textbf{Layer-4}} & \textit{\textbf{Layer-5}} & \textit{\textbf{Layer-6}} & \textit{\textbf{Layer-7}} \\
            \midrule
            0.18327915839536324 & 0.10736651162446896 & 0.1325381654743932 & 0.1517180627068569 \\
            \midrule
            \textit{\textbf{Layer-8}} & \textit{\textbf{Layer-9}} & \textit{\textbf{Layer-10}} & \textit{\textbf{Layer-11}} \\
            \midrule
            0.183028676828816 & 0.2298031295153093 & 0.04026268971247067 & 0.17620642501895548 \\
            \midrule
            \textit{\textbf{Layer-12}} & \textit{\textbf{Layer-13}} & \textit{\textbf{Layer-14}} & \textit{\textbf{Layer-15}} \\
            \midrule
            0.06616139460524989 & 0.07382263433361028 & 0.05470754299052264 & 0.1456038445898742 \\
            \midrule
            \textit{\textbf{Layer-16}} & \textit{\textbf{Layer-17}} & \textit{\textbf{Layer-18}} & \textit{\textbf{Layer-19}} \\
            \midrule
            0.015583363677548107 & 0.019572076662935905 & 0.08034528015969278 & 0.13836900028647664 \\
            \midrule
            \textit{\textbf{Layer-20}} & \textit{\textbf{Layer-21}} & \textit{\textbf{Layer-22}} & \textit{\textbf{Layer-23}} \\
            \midrule
            0.04670461556957088 & 0.02119316782149739 & 0.036045965083449205 & 0.04315987329920304 \\
            \midrule
            \textit{\textbf{Layer-24}} & \textit{\textbf{Layer-25}} & \textit{\textbf{Layer-26}} & \textit{\textbf{Layer-27}} \\
            \midrule
            0.05940025252392963 & 0.06281379336040903 & 0.07035992447924916 & 0.23457309192037937 \\
            \midrule
            \textit{\textbf{Layer-28}} & \textit{\textbf{Layer-29}} & \textit{\textbf{Layer-30}} & \textit{\textbf{Layer-31}} \\
            \midrule
            0.1578743191757248 & 0.23826246953092736 & 0.07079766552692811 & 0.89644860502407 \\
            \bottomrule
        \end{NiceTabular}
        }
	    \end{minipage}
\end{table}

\newpage
\section{Adopted Retrieval Heads}\label{appendix::rh}
The adopted retrieval heads and corresponding score are listed in triple (layer id, head id, score):

(5, 8, 0.21), (8, 1, 0.49), (10, 14, 0.45), (13, 6, 0.15), (14, 18, 0.15), (15, 30, 0.90), (16, 1, 0.50), (17, 29, 0.12), (19, 3, 0.30), (20, 14, 0.44), (22, 14, 0.33), (24, 27, 0.46), (26, 15, 0.14), (27, 7, 0.30)

\section{Supplementary Contents on Semantic Span Division and NMS Acceleration}\label{appendix::span_division_nms}

\begin{figure}[H]
    \subfloat[For-Loop]{%
        \includegraphics[width=.48\linewidth]{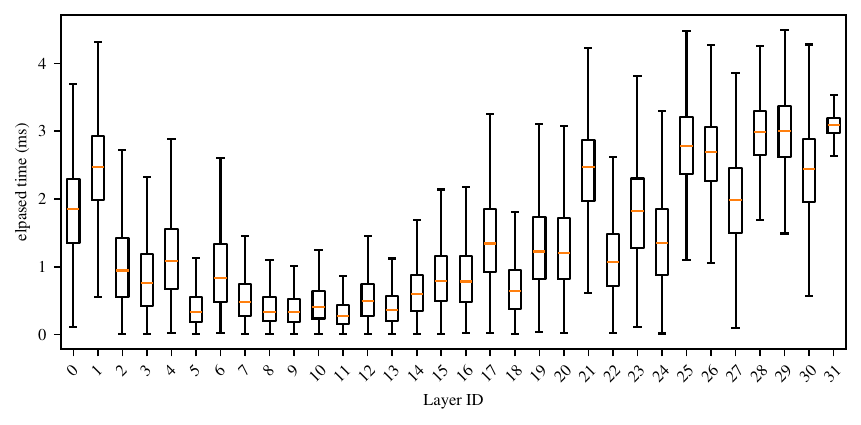}%
        \label{fig:span_generation_time_for_loop}%
    }\hfill
    \subfloat[Non-Maximum Suppression]{%
        \includegraphics[width=.48\linewidth]{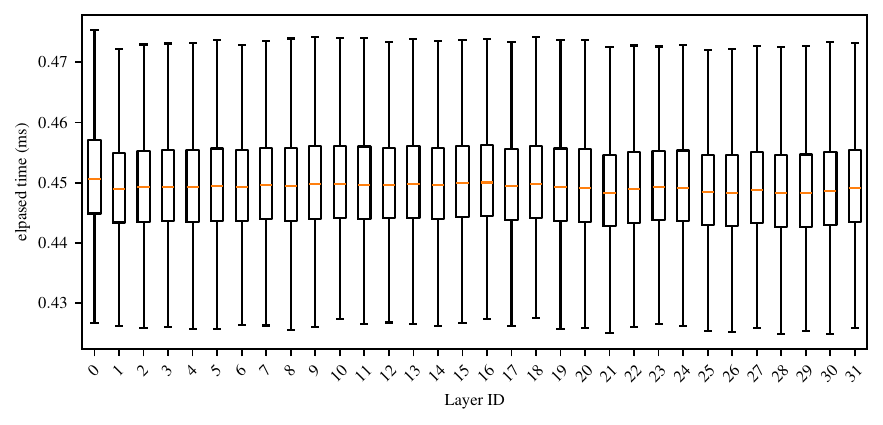}%
        \label{fig:span_generation_time_nms}%
    }
    \caption{Semantic Span Generation Time based on Different Implementations.}
    \label{fig:span_generation_time}
\end{figure}

In contrary to the for-loop implementation for semantic span division, utilizing the NMS operator from the torchvision library facilitates faster and steady semantic span generation, as shown in Figure \ref{fig:span_generation_time}.

\begin{figure}[htbp]
    \subfloat[\textbf{\textit{$\varphi=0.01$}}]{%
        \includegraphics[width=.24\linewidth]{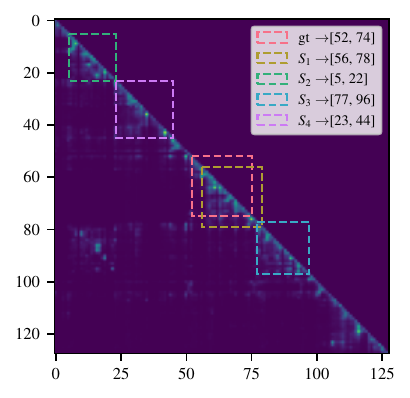}%
        \label{fig:semantic_span_division_iou_0.01}%
    }\hfill
    \subfloat[\textbf{\textit{$\varphi=0.1$}}]{%
        \includegraphics[width=.24\linewidth]{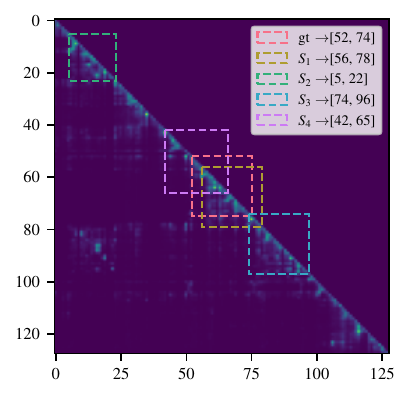}%
        \label{fig:semantic_span_division_iou_0.1}%
    }\hfill
    \subfloat[\textbf{\textit{$\varphi=0.4$}}]{%
        \includegraphics[width=.24\linewidth]{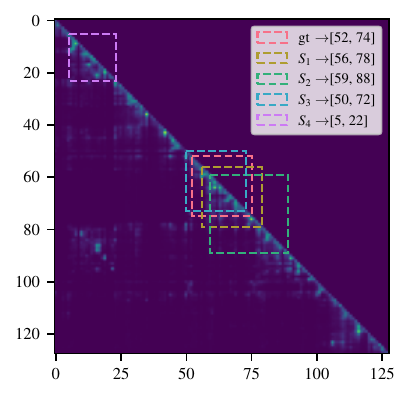}%
        \label{fig:semantic_span_division_iou_0.4}%
    }\hfill
    \subfloat[\textbf{\textit{$\varphi=0.7$}}]{%
        \includegraphics[width=.24\linewidth]{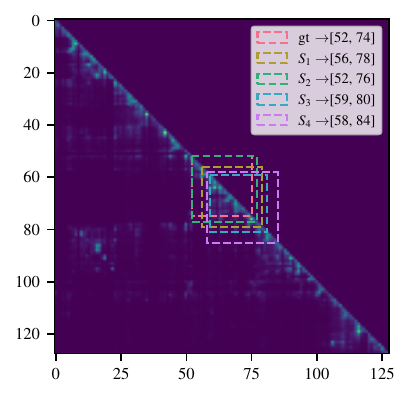}%
        \label{fig:semantic_span_division_iou_0.7}%
    }\\
    \caption{Semantic span division results under different Intersection over Union Threshold $\varphi$.}
    \label{fig:semantic_span_division_different_iou_thresholds}
\end{figure}

To illustrate the influence of Intersection over Union Threshold $\varphi$, we additionally visualize the semantic span division results under different $\varphi$ in Figure \ref{fig:semantic_span_division_different_iou_thresholds}, with the attention map threshold $\theta$ fixed at $90^{\text{th}} \text{ percentile}$. It's observed that lower $\varphi$ values result in more dispersed semantic spans, as depicted in Figure \ref{fig:semantic_span_division_iou_0.01}. As $\varphi$ increases, the divided semantic spans overlap in nearby regions in Figure \ref{fig:semantic_span_division_iou_0.7}. In summary, lower $\varphi$ values encourage semantic span discovery, increasing the likelihood of evidence retention and the risk of noise inclusion. Conversely, higher $\varphi$ values can boost algorithm confidence in semantic span division results, but potentially losing evidences.

\begin{figure}[htbp]
    \subfloat[\textbf{\textit{Layer-0}}]{%
        \includegraphics[width=.24\linewidth]{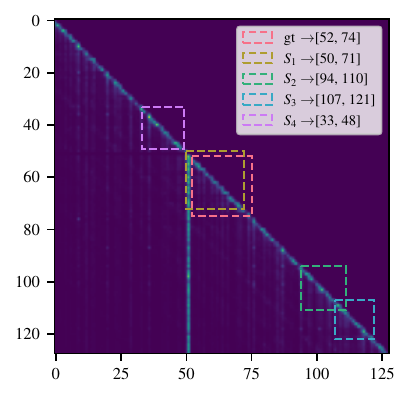}%
        \label{fig:semantic_span_division_layer_0}%
    }\hfill
    \subfloat[\textbf{\textit{Layer-1}}]{%
        \includegraphics[width=.24\linewidth]{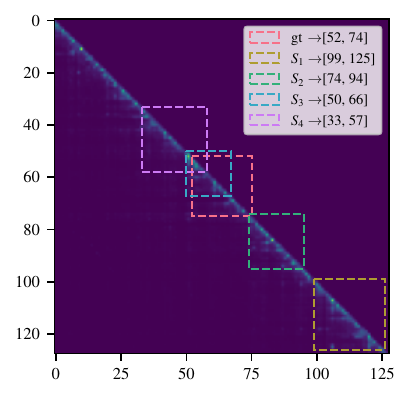}%
        \label{fig:semantic_span_division_layer_1}%
    }\hfill
    \subfloat[\textbf{\textit{Layer-2}}]{%
        \includegraphics[width=.24\linewidth]{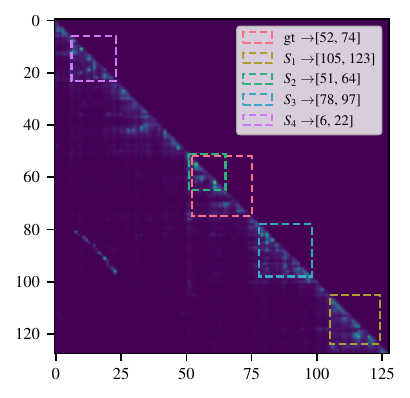}%
        \label{fig:semantic_span_division_layer_2}%
    }\hfill
    \subfloat[\textbf{\textit{Layer-3}}]{%
        \includegraphics[width=.24\linewidth]{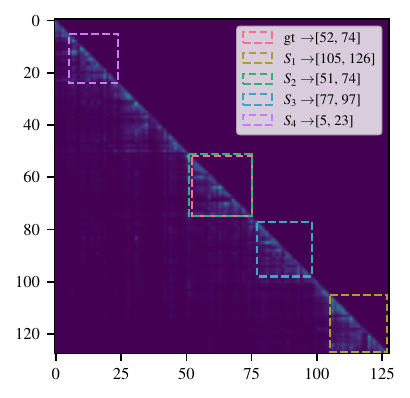}%
        \label{fig:semantic_span_division_layer_3}%
    }\\
    \subfloat[\textbf{\textit{Layer-4}}]{%
        \includegraphics[width=.24\linewidth]{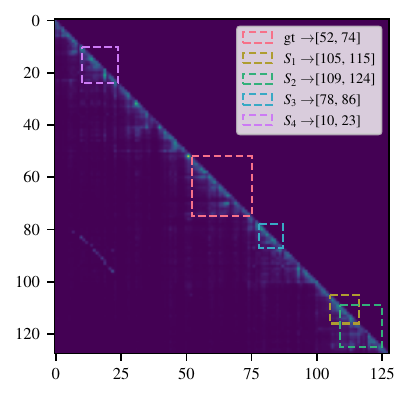}%
        \label{fig:semantic_span_division_layer_4}%
    }\hfill
    \subfloat[\textbf{\textit{Layer-5}}]{%
        \includegraphics[width=.24\linewidth]{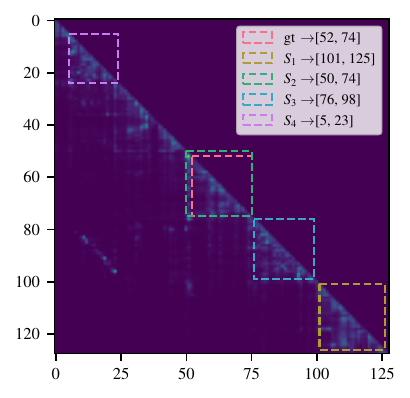}%
        \label{fig:semantic_span_division_layer_5}%
    }\hfill
    \subfloat[\textbf{\textit{Layer-6}}]{%
        \includegraphics[width=.24\linewidth]{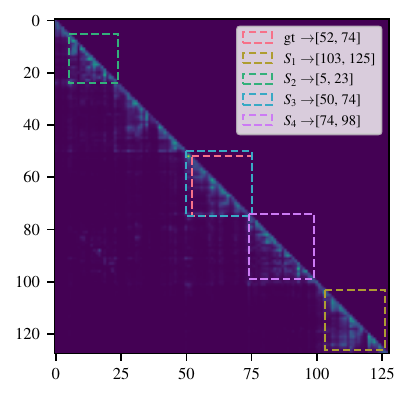}%
        \label{fig:semantic_span_division_layer_6}%
    }\hfill
    \subfloat[\textbf{\textit{Layer-7}}]{%
        \includegraphics[width=.24\linewidth]{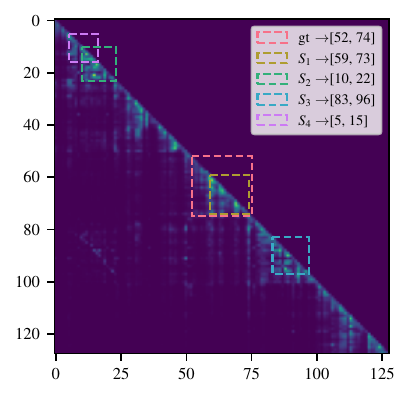}%
        \label{fig:semantic_span_division_layer_7}%
    }\\
    \subfloat[\textbf{\textit{Layer-8}}]{%
        \includegraphics[width=.24\linewidth]{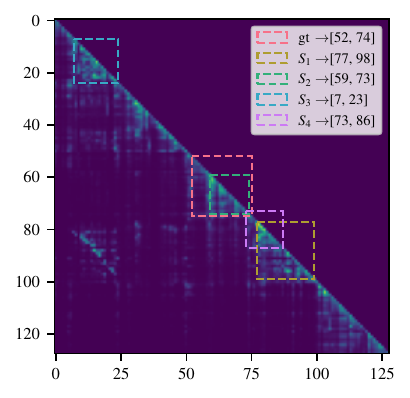}%
        \label{fig:semantic_span_division_layer_8}%
    }\hfill
    \subfloat[\textbf{\textit{Layer-9}}]{%
        \includegraphics[width=.24\linewidth]{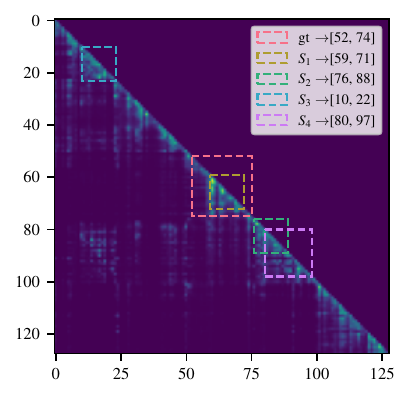}%
        \label{fig:semantic_span_division_layer_9}%
    }\hfill
    \subfloat[\textbf{\textit{Layer-10}}]{%
        \includegraphics[width=.24\linewidth]{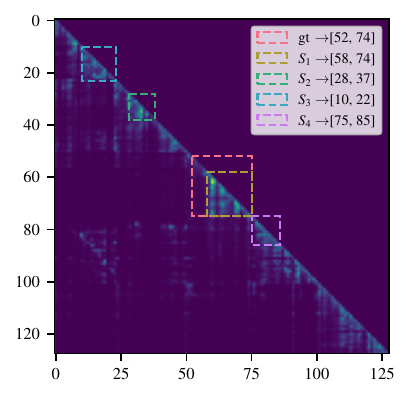}%
        \label{fig:semantic_span_division_layer_10}%
    }\hfill
    \subfloat[\textbf{\textit{Layer-11}}]{%
        \includegraphics[width=.24\linewidth]{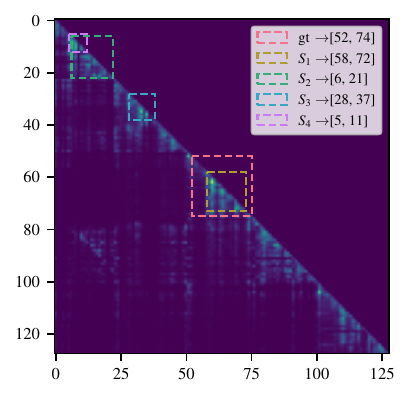}%
        \label{fig:semantic_span_division_layer_11}%
    }\\
    \subfloat[\textbf{\textit{Layer-12}}]{%
        \includegraphics[width=.24\linewidth]{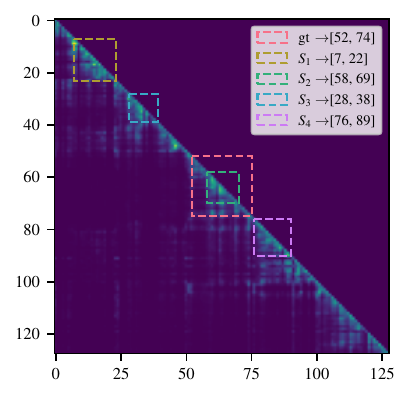}%
        \label{fig:semantic_span_division_layer_12}%
    }\hfill
    \subfloat[\textbf{\textit{Layer-13}}]{%
        \includegraphics[width=.24\linewidth]{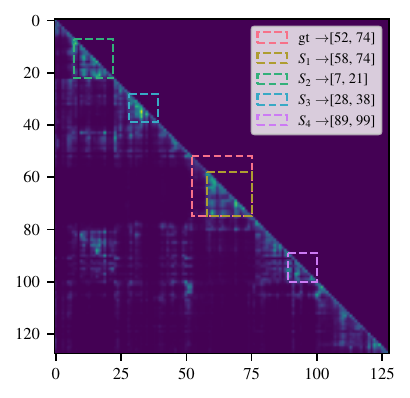}%
        \label{fig:semantic_span_division_layer_13}%
    }\hfill
    \subfloat[\textbf{\textit{Layer-14}}]{%
        \includegraphics[width=.24\linewidth]{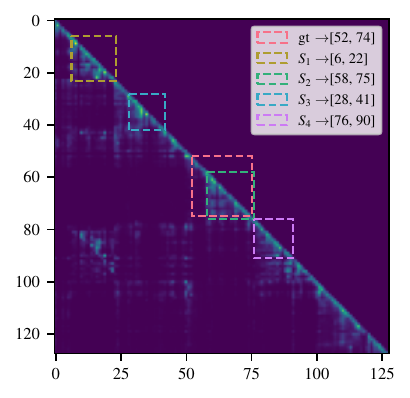}%
        \label{fig:semantic_span_division_layer_14}%
    }\hfill
    \subfloat[\textbf{\textit{Layer-15}}]{%
        \includegraphics[width=.24\linewidth]{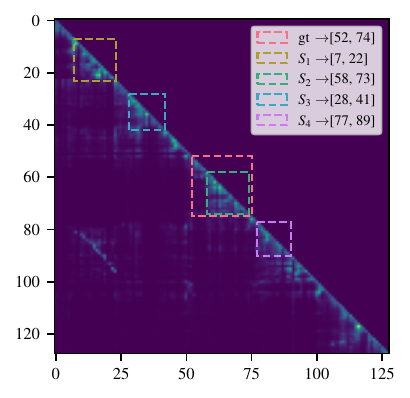}%
        \label{fig:semantic_span_division_layer_15}%
    }\\
    \caption{Demonstrated Semantic Span Division Results for Layers 0-15.}
    \label{fig:semantic_span_division_layers_0_15}
\end{figure}

\begin{figure}[htbp]
    \subfloat[\textbf{\textit{Layer-16}}]{%
        \includegraphics[width=.24\linewidth]{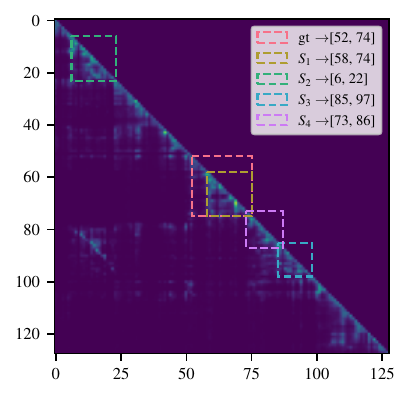}%
        \label{fig:semantic_span_division_layer_16}%
    }\hfill
    \subfloat[\textbf{\textit{Layer-17}}]{%
        \includegraphics[width=.24\linewidth]{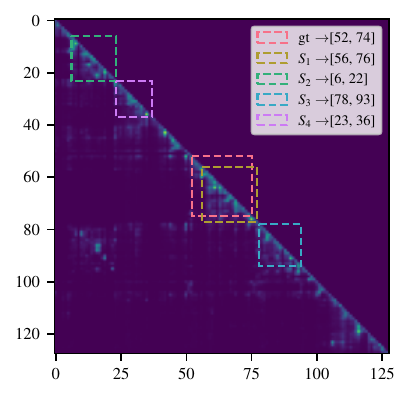}%
        \label{fig:semantic_span_division_layer_17}%
    }\hfill
    \subfloat[\textbf{\textit{Layer-18}}]{%
        \includegraphics[width=.24\linewidth]{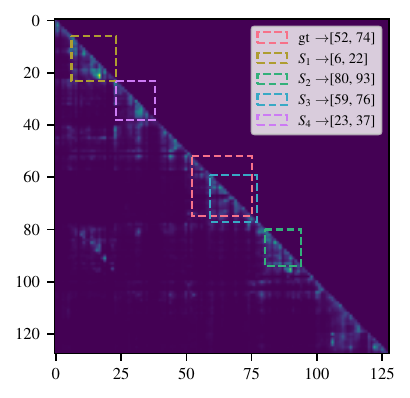}%
        \label{fig:semantic_span_division_layer_18}%
    }\hfill
    \subfloat[\textbf{\textit{Layer-19}}]{%
        \includegraphics[width=.24\linewidth]{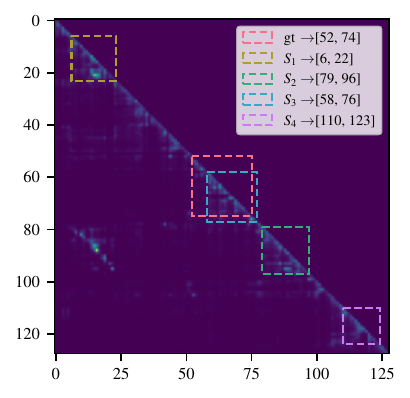}%
        \label{fig:semantic_span_division_layer_19}%
    }\\
    \subfloat[\textbf{\textit{Layer-20}}]{%
        \includegraphics[width=.24\linewidth]{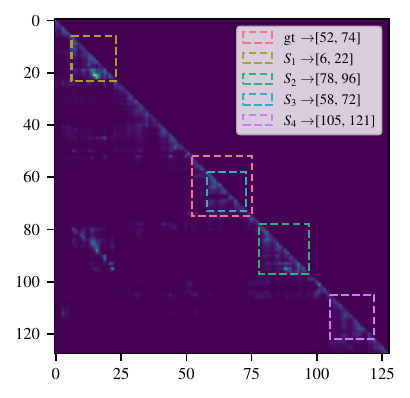}%
        \label{fig:semantic_span_division_layer_20}%
    }\hfill
    \subfloat[\textbf{\textit{Layer-21}}]{%
        \includegraphics[width=.24\linewidth]{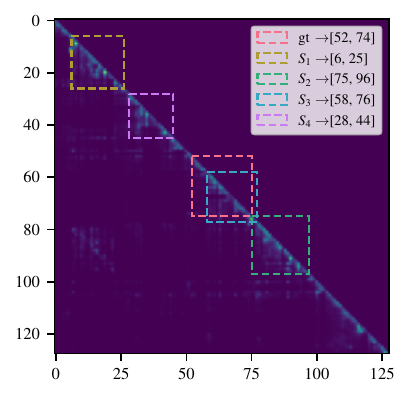}%
        \label{fig:semantic_span_division_layer_21}%
    }\hfill
    \subfloat[\textbf{\textit{Layer-22}}]{%
        \includegraphics[width=.24\linewidth]{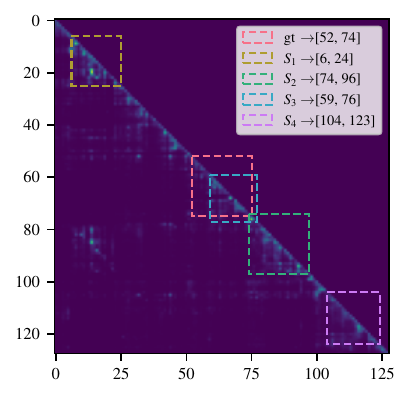}%
        \label{fig:semantic_span_division_layer_22}%
    }\hfill
    \subfloat[\textbf{\textit{Layer-23}}]{%
        \includegraphics[width=.24\linewidth]{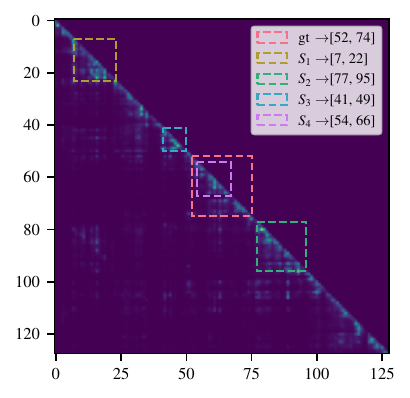}%
        \label{fig:semantic_span_division_layer_23}%
    }\\
    \subfloat[\textbf{\textit{Layer-24}}]{%
        \includegraphics[width=.24\linewidth]{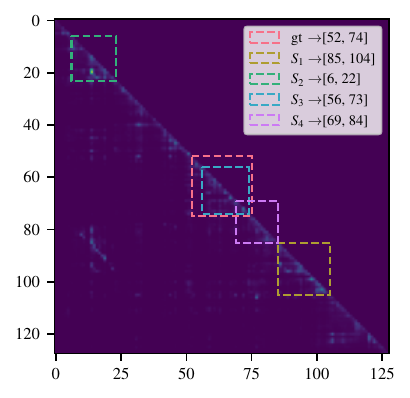}%
        \label{fig:semantic_span_division_layer_24}%
    }\hfill
    \subfloat[\textbf{\textit{Layer-25}}]{%
        \includegraphics[width=.24\linewidth]{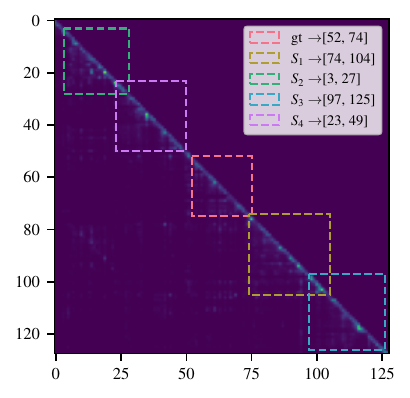}%
        \label{fig:semantic_span_division_layer_25}%
    }\hfill
    \subfloat[\textbf{\textit{Layer-26}}]{%
        \includegraphics[width=.24\linewidth]{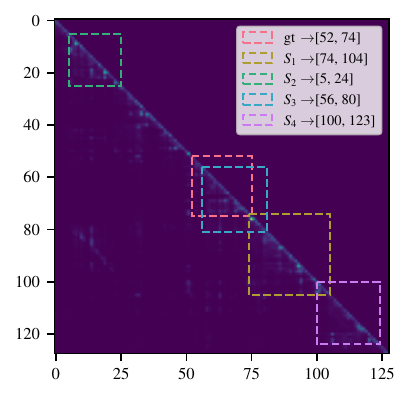}%
        \label{fig:semantic_span_division_layer_26}%
    }\hfill
    \subfloat[\textbf{\textit{Layer-27}}]{%
        \includegraphics[width=.24\linewidth]{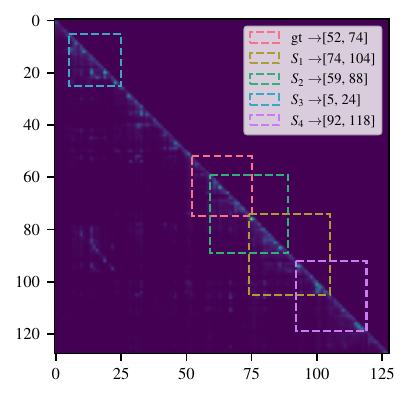}%
        \label{fig:semantic_span_division_layer_27}%
    }\\
    \subfloat[\textbf{\textit{Layer-28}}]{%
        \includegraphics[width=.24\linewidth]{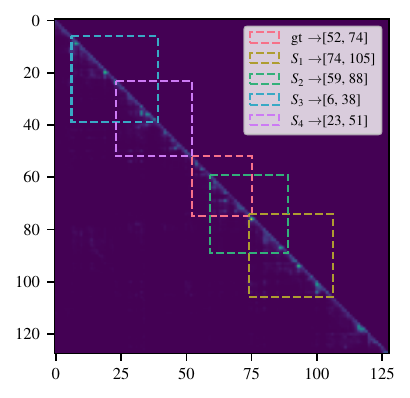}%
        \label{fig:semantic_span_division_layer_28}%
    }\hfill
    \subfloat[\textbf{\textit{Layer-29}}]{%
        \includegraphics[width=.24\linewidth]{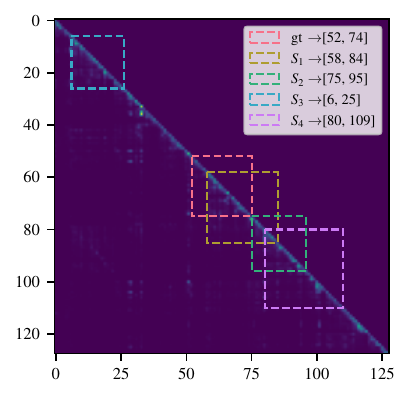}%
        \label{fig:semantic_span_division_layer_29}%
    }\hfill
    \subfloat[\textbf{\textit{Layer-30}}]{%
        \includegraphics[width=.24\linewidth]{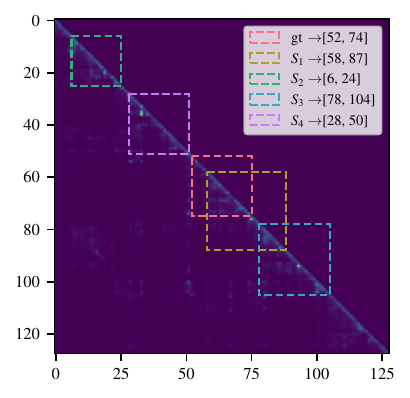}%
        \label{fig:semantic_span_division_layer_30}%
    }\hfill
    \subfloat[\textbf{\textit{Layer-31}}]{%
        \includegraphics[width=.24\linewidth]{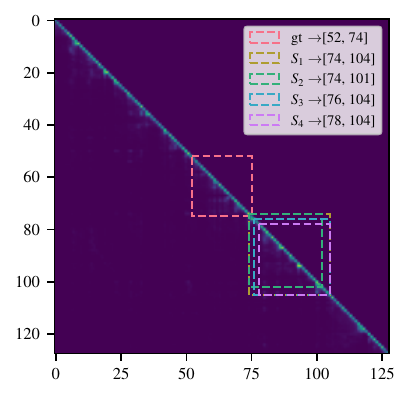}%
        \label{fig:semantic_span_division_layer_31}%
    }\\
    \caption{Demonstrated Semantic Span Division Results for Layers 16-31.}
    \label{fig:semantic_span_division_layers_16_31}
\end{figure}

We visualize the semantic span division results for every individual attention layer from one block containing the needle in Figure \ref{fig:semantic_span_division_layers_0_15} and \ref{fig:semantic_span_division_layers_16_31}. In most cases, our proposed strategy can find at lease one span overlapping with the ground truth needle position. What's more, there are situations where exist a semantic span that overlaps with the needle accurately.

\section{NMS Input Filtration for Efficient Span Selection}\label{appendix::span_selection_alg}
After Eq.\ref{eq:sxy}, each point $(i,j)$ in the attention map is associated with a token span interval $[j, i]$ and triangular cumulative score. The selected points are used to construct the corresponding box for Non-Maximum Suppression (NMS) calculations. To decrease the computational complexity of the NMS process, we can optionally limit the number of points involved in the calculation. Here, we opt to select the highest scoring points on each anti-diagonal as our filtering strategy. The approach linearly correlates the complexity of the NMS process to the number of tokens in the attention map, and makes the remaining points more likely to correspond to the midpoint of the semantic spans.

Let $[p, q]$ be the token interval covering the evidence. Ideally, the cumulative triangular score for $[(p+q)//2, (p+q)//2]$, $\dots$, $[p, q]$, $\dots$, $[1, N]$ should show a trend of first increasing and then decreasing. The segment from $[(p+q)//2, (p+q)//2]$ to $[p, q]$ is located in the same semantic span, so the score gradually increases; the segment from $[p, q]$ to $[1, N]$ contains tokens outside of the semantic span, under the operation of minus attention map threshold $\theta$, the score of segments gradually decreases.

\section{Detailed Comparison with MInference on RULER Benchmark}\label{appendix::detailed_comparison_with_MInference_on_RULER}
\begin{table*}[htbp]
\caption{MInference versus Ltri-LLM on RUELR Benchmark. The scores of MInference and Ltri-LLM are listed before and after the slash (/), respectively. The absolute performance difference of Ltri-LLM relative to MInference is indicated in parentheses.}
\label{tab:ruler-comparison}
\begin{center}
\begin{small}
\resizebox{\textwidth}{!}{
    \begin{NiceTabular}{@{}cccccccc@{}}
    \CodeBefore
        \rowcolor{lightgray!20}{2}
        \rowcolor{lightgray!20}{4}
        \rowcolor{lightgray!20}{6}
        \rowcolor{lightgray!20}{9}
        \rowcolor{lightgray!20}{11}
        \rowcolor{lightgray!20}{13}
    \Body
        \toprule
        SEQ\_LENGTH & niah\_single\_1 & niah\_single\_2 & niah\_single\_3 & niah\_multikey\_1 & niah\_multikey\_2 & niah\_multikey\_3 & niah\_multivalue \\
        \midrule
        4096 & 100.0/100.0(+0.0) & 100.0/100.0(+0.0) & 100.0/100.0(+0.0) & 100.0/100.0(+0.0) & 100.0/100.0(+0.0) & 100.0/100.0(+0.0) & 92.0/92.0(+0.0) \\
        8192 & 100.0/100.0(+0.0) & 100.0/100.0(+0.0) & 100.0/100.0(+0.0) & 100.0/100.0(+0.0) & 100.0/100.0(+0.0) & 100.0/100.0(+0.0) & 87.0/79.0(-8.0) \\
        16384 & 100.0/100.0(+0.0) & 100.0/100.0(+0.0) & 100.0/100.0(+0.0) & 96.0/95.7(-0.3) & 100.0/85.7(-14.3) & 100.0/81.2(-18.8) & 98.0/78.0(-20.0) \\
        32768 & 100.0/100.0(+0.0) & 100.0/100.0(+0.0) & 100.0/100.0(+0.0) & 100.0/95.8(-4.2) & 96.0/84.0(-12.0) & 96.0/63.2(-32.8) & 87.0/78.0(-9.0) \\
        65536 & 100.0/100.0(+0.0) & 100.0/100.0(+0.0) & 100.0/100.0(+0.0) & 100.0/72.7(-27.3) & 100.0/85.0(-15.0) & 92.0/50.0(-42.0) & 98.0/86.0(-12.0) \\
        131072 & 100.0/100.0(+0.0) & 100.0/94.1(-5.9) & 100.0/100.0(+0.0) & 100.0/82.6(-17.4) & 100.0/77.3(-22.7) & 52.0/85.7(+33.7) & 95.0/88.0(-7.0) \\
        \midrule
        SEQ\_LENGTH & niah\_multiquery & vt & cwe & fwe & qa\_1 & qa\_2 & AVG \\
        \midrule
        4096 & 100.0/100.0(+0.0) & 100.0/100.0(+0.0) & 93.2/92.8(-0.4) & 93.3/93.3(+0.0) & 72.0/72.0(+0.0) & 56.0/56.0(+0.0) & 92.8/92.8(-0.0) \\
        8192 & 100.0/98.0(-2.0) & 93.6/75.2(-18.4) & 73.2/73.2(+0.0) & 84.0/82.7(-1.3) & 56.0/56.0(+0.0) & 40.0/36.0(-4.0) & 87.2/84.6(-2.6) \\
        16384 & 100.0/93.0(-7.0) & 95.2/72.0(-23.2) & 44.4/47.6(+3.2) & 89.3/93.3(+4.0) & 72.0/64.0(-8.0) & 40.0/40.0(+0.0) & 87.3/80.8(-6.5) \\
        32768 & 100.0/87.0(-13.0) & 90.4/66.4(-24.0) & 4.0/24.0(+20.0) & 84.0/88.0(+4.0) & 64.0/40.0(-24.0) & 48.0/28.0(-20.0) & 82.3/73.4(-8.8) \\
        65536 & 99.0/95.0(-4.0) & 88.0/69.6(-18.4) & 0.8/2.8(+2.0) & 86.7/85.3(-1.3) & 68.0/48.0(-20.0) & 44.0/16.0(-28.0) & 82.8/70.0(-12.8) \\
        131072 & 98.0/97.0(-1.0) & 81.6/62.4(-19.2) & 0.8/1.6(+0.8) & 74.7/74.7(+0.0) & 56.0/36.0(-20.0) & 44.0/32.0(-12.0) & 77.1/71.6(-5.4) \\
        \bottomrule
    \end{NiceTabular}
}
\end{small}
\end{center}
\vskip -0.2in
\end{table*}

\begin{figure}[H]
    \centering
    \includegraphics[width=0.95\textwidth]{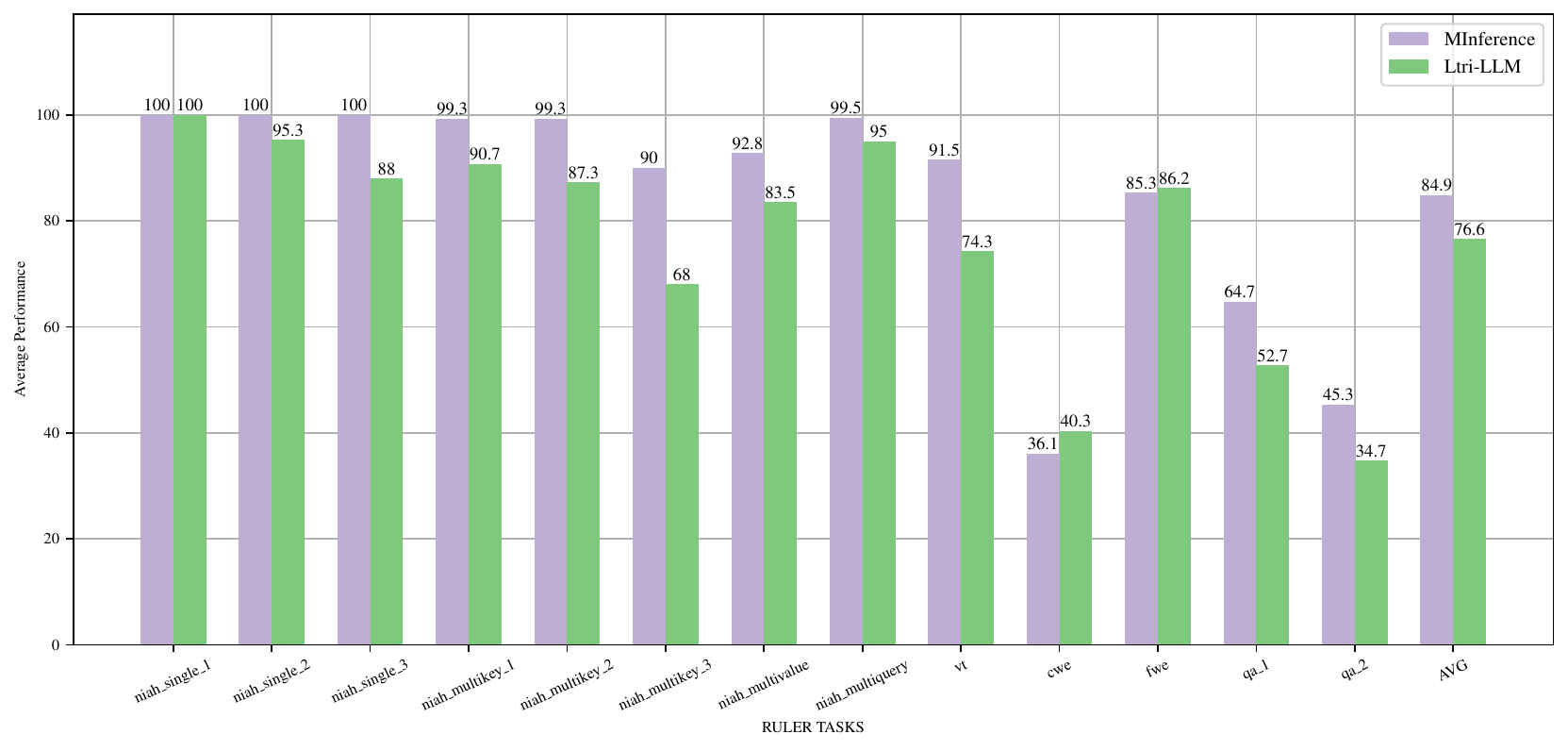}
    \caption{Performance Comparison between MInference and Ltri-LLM on RULER tasks. The average performance ranging from 4K to 128K of each task is reported. For niah\_single\_1,  niah\_single\_2,  niah\_single\_3,  niah\_multikey\_1,  niah\_multikey\_2,  niah\_multikey\_3, Ltri-LLM reports cases whose evidences locate within a block.}
    \label{fig:detailed_comparison_with_MInference_on_RULER}
\end{figure}

\end{document}